\documentclass{article}
\usepackage[accepted]{icml2022} 
\usepackage{algorithmic}
\usepackage{algorithm}
\usepackage{amsmath}
\usepackage{microtype}
\usepackage{booktabs}
\usepackage{multirow} 
\usepackage{enumitem}
\usepackage{ textcomp }
\usepackage{pifont}
\newcommand{\xmark}{\ding{54}}%
\usepackage{mathtools,  amsmath }

\usepackage[toc,page]{appendix}
\usepackage{amssymb}
\usepackage{amsthm}
\renewcommand{\emph}[1]{\textit{#1}}
\DeclareTextFontCommand{\emph}{\textit}

\usepackage{hyperref}
\definecolor{frenchblue}{rgb}{0.0, 0.45, 0.73}
\definecolor{orange}{rgb}{0.988, 0.545, 0.231}
\definecolor{bl}{rgb}{0.145 0.424 0.671}
\hypersetup{
  colorlinks,
  citecolor=frenchblue,
  linkcolor=frenchblue,
  urlcolor=frenchblue}

\usepackage{tcolorbox}
\definecolor{lightblue}{HTML}{84C7F9}
\definecolor{pltorange}{HTML}{f86910}
\definecolor{pltblue}{HTML}{1b62a5}
\definecolor{bleudefrance}{rgb}{0.19, 0.55, 0.91}

\definecolor{lighterblue}{HTML}{D4ECFF}
\newtcolorbox{mybox}{colback=lighterblue,colframe=lightblue}
\usepackage{siunitx}
\usepackage{float}
\usepackage{placeins}
\usepackage{graphicx}
\usepackage{xcolor}
\usepackage{mathtools}
\usepackage{makecell}

\definecolor{frenchblue}{rgb}{0.0, 0.45, 0.73}

\definecolor{Red}{rgb}{0.8, 0.0, 0.0}
\definecolor{Violet}{rgb}{0.16, 0.32, 0.75}
\definecolor{trolleygrey}{rgb}{0.5, 0.5, 0.5}
\definecolor{trolleygrey1}{rgb}{0.2, 0.2, 0.2}
\definecolor{dpgreen}{rgb}{0.01, 0.75, 0.24}
\definecolor{darkpink}{rgb}{0.91, 0.33, 0.5}
\definecolor{alizarin}{rgb}{0.82, 0.1, 0.26}
\definecolor{americanrose}{rgb}{1.0, 0.01, 0.24}

\definecolor{pistachio}{rgb}{0.58, 0.77, 0.45}

\def\eqref#1{equation~\ref{#1}}

\def\1{\bm{1}}

\DeclareMathAlphabet{\mathsfit}{\encodingdefault}{\sfdefault}{m}{sl}
\SetMathAlphabet{\mathsfit}{bold}{\encodingdefault}{\sfdefault}{bx}{n}

\DeclareMathOperator{\sign}{sign}

\newcommand{\method}{AdaTQC }
\newcommand{\maxmin}{MMQL}

\DeclareMathOperator{\rollout}{\rho}
\DeclareMathOperator{\bias}{bias}
\DeclareMathOperator{\aggbias}{agg bias}
\usepackage{amsfonts}
\usepackage{ulem}
\usepackage{utfsym}
\usepackage{pifont}
\newcommand{\walkersym}{\usym{1F6B6}}
\newcommand{\coptersym}{\ding{40}}
\usetikzlibrary{calc, arrows, decorations.markings}

\newcommand\wavyaggbias{\stackrel{\mathclap{\normalfont\mbox{\uwave{\quad\quad\quad}}}}{\aggbias}}

\usepackage{subcaption}
\usepackage{graphicx}

\begin{document}

\twocolumn[
\icmltitle{Automating Control of Overestimation Bias for Reinforcement Learning}



\icmlsetsymbol{equal}{*}

\begin{icmlauthorlist}
\icmlauthor{Arsenii Kuznetsov}{equal,saic}
\icmlauthor{Alexander Grishin}{equal,airi,hse}
\icmlauthor{Artem Tsypin}{airi}

\icmlauthor{Arsenii Ashukha}{saic,hse}
\icmlauthor{Artur Kadurin}{airi}
\icmlauthor{Dmitry Vetrov}{airi,hse}
\end{icmlauthorlist}

\icmlaffiliation{saic}{Samsung AI Center, Moscow}
\icmlaffiliation{airi}{Artificial Intelligence Research Institute, Moscow}
\icmlaffiliation{hse}{National Research University Higher School of Economics, Moscow}

\icmlcorrespondingauthor{Arsenii Kuznetsov}{brickerino@gmail.com}
\icmlcorrespondingauthor{Alexander Grishin}{grishin.alexgri@yandex.ru}

\icmlkeywords{Machine Learning, ICML}

\vskip 0.3in
]

\printAffiliationsAndNotice{\icmlEqualContribution}
\begin{abstract}
Overestimation bias control techniques are used by the majority of high-performing off-policy reinforcement learning algorithms.
However, most of these techniques rely on a pre-defined bias correction policies that are either not flexible enough or require environment-specific tuning of hyperparameter.
In this work, we present a general data-driven approach for automatic selection of bias control hyperparameters.
We demonstrate its effectiveness on three algorithms: Truncated Quantile Critics, Weighted Delayed DDPG and Maxmin Q-learning. 
The proposed technique eliminates the need for an extensive hyperparameter search. 
We show that it leads to the significant reduction of the actual number of interactions while preserving the performance.
\end{abstract}

\section{Introduction}
An accurate estimation of a state-action value function~(Q-function) is the cornerstone of reinforcement learning.
Overestimation bias is a source of error in value function estimates~\citep{thrun1993issues, hasselt2010double}. 
It emerges from inaccurate value estimates in temporal-difference learning, where an imprecise sample-based maximization leads to accumulation of  errors~\citep{fujimoto2018addressing}.
As a consequence of the bias, sub-optimal actions may receive high-value estimates. This slows down the training and leads to inferior results.

The desire for bias correction has inspired a number of high-performing reinforcement learning algorithms \citep{fujimoto2018addressing,lan2020maxmin, kuznetsov2020controlling}.
However, to adjust the magnitude of control, these algorithms require outer-loop hyperparameter tuning, effectively, worsening the interaction efficiency of reinforcement learning algorithms. Yet, the bias remains not fully eliminated~\citep{he2020reducing}.

\begin{figure}[t!]
\centering
\includegraphics[width=0.48\textwidth]{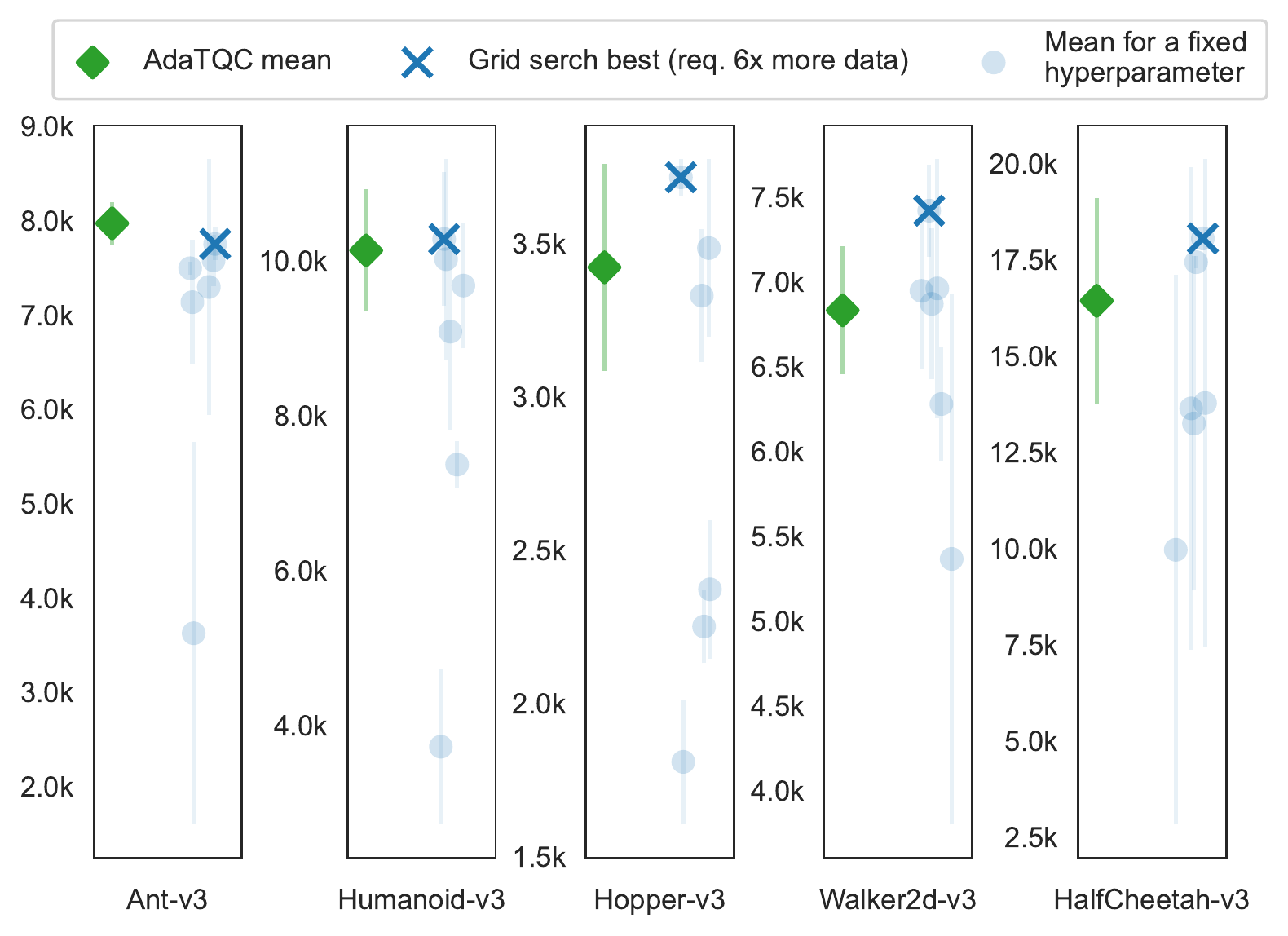}
\caption{The results of the proposed AdaTQC---an adaptive version of TQC---on MuJoCo. AdaTQC achieves the level of performance of a \textit{Grid search} while requiring only a single run. Lines represent standard deviations across seeds.}
\label{fig:ada_tqc}
\end{figure}

In this work, we show that the bias can be directly reduced on the fly during a single run.
We propose a simple technique that allows for a correction based on the estimated bias.
We apply the proposed adaptation technique to the Truncated Quantile Critics~(TQC)~\citep{kuznetsov2020controlling}, Weighted Delayed Deep Deterministic Policy Gradient~(WD3)~\citep{lan2020maxmin}, and Maxmin Q-learning~(MMQL)~\citep{lan2020maxmin}.
These algorithms are both sensitive to hyperparameters and require an environment-specific tuning of hyperparameters~(reported in the original works and in Appendix~\ref{app:heatmap}).
The on-the-fly adaptation of the bias control allows to avoid a costly outer loop hyperparameter search.
This reduces an actual number of required environment interactions while preserving the performance (e.g., Figure \ref{fig:ada_tqc}).

Our contributions can be summarized as follows.
\begin{itemize}[noitemsep,topsep=0pt]
    \setlength\itemsep{0em}
    \item[$a)$] We propose a simple method that automatically selects a magnitude of an overestimation control on the fly.
    \item[$b)$] We show a significant reduction of a number of interactions while preserving the performance for three bias control algorithms: TQC, WD3 and \maxmin.
    \item[$c)$] We investigate the proposed estimate of bias and its connection to the true bias.
\end{itemize}

\section{Background}
\begin{table}[t!]
\begin{center}
\resizebox{0.48\textwidth}{!}{%
\begin{tabular}{l l}
\toprule
\bf Notation & \bf Description \\
\midrule
\bf Q-functions\\
~~~~$Q^\pi$ & State-action value function for policy $\pi$\\
~~~~$\hat{Q}_\psi$ & Approximation to state-action value function\\
~~~~$N$ & Number of value function approximations\\
~~~~$\hat{Q}_{\psi_n}$ & $n$-th approximation out of $N$\\
~~~~$\hat{Q}_{t}$ & approximation at step $t$ of optimization\\
~~~~$\tilde{Q}$ & Estimate of state-action value function\\
\midrule
\bf Distributional RL\\
~~~~$Z^\pi$ & Distributional state-action value function\\
\multirow{2}{*}{~~~~$\hat{Z}_{\psi_n}$} &  Approximation to distributional value function\\
& returning $M$ scalars\\
\multirow{2}{*}{~~~~$\hat{Z}^m_{\psi_n}$} & $m$-th atom of $n$-th approximation \\
& to distriubtional state-action value function \\
~~~~$x_{(i)}$ & $i$-th element of $x$ sorted in ascending order\\
\midrule
\bf Bias control\\
~~~~$\bias(\hat{Q}, \pi, s, a)$ &  Signed estimation error; i.e., $\hat{Q}(s, a) - Q^{\pi}(s,a)$\\
~~~~$d_\pi(s), d_\pi(s, a)$ &  State(-action) visitation distribution\\
~~~~$\aggbias(\hat{Q}, \pi)$ &  Aggregated bias; i.e., $\mathbb{E}_{s,a \sim d_\pi(s, a)} \bias(\hat{Q}, \pi, s, a)$\\
~~~~$k$ &  Number of future rewards used in return estimate\\
~~~~\multirow{2}{*}{$\rollout$} &  Rollout, containing $k$ future steps \\
&(or less if \textit{done} achieved in less than $k$ steps)\\
~~~~$\mathcal{B}$ &  Set of rollouts (potentially intersecting)\\
~~~~$\wavyaggbias(\hat{Q}, \{\rollout_l\}_{l=1}^L)$ &  Estimate of aggregated bias based on $L$ rollouts\\
~~~~$\eta$ & Hyperparameter controlling overestimation bias \\

\bottomrule
\end{tabular}
}
\end{center}
\caption{Notation used in the paper.}
\label{tab:notation}
\end{table}

In this section, we overview three RL algorithms with an internal mechanisms of overestimation bias control.
All the necessary notation is described in Table~\ref{tab:notation}.

\subsection{Truncated Quantile Critics (TQC)}
TQC \cite{kuznetsov2020controlling} is a continuous control algorithm. 
It  uses multiple approximations $\hat{Z}_{\psi_n}(s,a)$ of the distributional state-action value function $Z^\pi(s,a)$, that are trained with quantile regression loss \cite{dabney2018distributional}.
Each approximation $\hat{Z}_{\psi_n}(s,a)$ outputs a set of $M$ atoms $\{\hat{Z}^m_{\psi_n}(s,a)\}_{m=1}^M$, each representing a quantile of the distribution over future discounted (entropy adjusted) returns~$Z^\pi(s,a)$.
In the TD-learning target TQC truncates  discrete number $\eta^{\text{\sl TQC}}$ of topmost atoms in order to cope with the overestimation bias.  

Each network $\hat{Z}_{\psi_n}$ learns $M$ quantiles of a distribution whose samples are given by 
\begin{equation}
    \Big\{ r(s, a) + 
    \gamma \big(z_{(i)}(s', a') - \alpha \log \pi (a'|s')\big)  \Big\}_{i=1}^{NM-\eta^{\text{\sl TQC}}},
\end{equation}
where $z_{(i)}(s', a')$ is $i$-th element in sorted sequence of all atoms from all approximations $\{\hat{Z}^m_{\psi_n}(s',a')\}_{n,m=1}^{N, M}$. 

As reported in the original paper TQC is sensitive to the amount of truncated atoms~$\eta^{\text{\sl TQC}}$ which has different optimal values for different environments.  

\subsection{Weighted Delayed DDPG (WD3)}
WD3 \cite{he2020wd3} is a continuous control algorithm that is based on \textit{twin delayed DDPG} algorithm \cite{fujimoto2018addressing}.
It uses weighted sum of the minimum and the average of two value functions approximations ${\hat{Q}_{\psi_i}(s', a')}_{i=1,2}$ to avoid possible underestimation bias of pure minimum. Specifically, 
it leads to the following temporal difference target
\begin{equation}
\begin{split}
    r(s, a) &+ \gamma \Big( \eta^{\text{\sl WD3}} \min_{i=1,2}\hat{Q}_{\psi_i}(s', a')
    \\
    &+ (1 - \eta^{\text{\sl WD3}}) \underset{i=1,2}{\mathrm{avg}}\hat{Q}_{\psi_i}(s', a') \Big),
\end{split}
\end{equation}
where $\min(\cdot)$ and $\mathrm{avg}(\cdot)$ are minimum and average functions respectively. As we show in Appendix~\ref{app:heatmap} a weight $\eta^{\text{\sl WD3}}$ requires per-environment tuning.
\subsection{Maxmin Q-learning (\maxmin)}
Maxmin Q-learning \cite{lan2020maxmin} is a discreet control algorithm. 
In contrast to conventional deep Q-learning \cite{mnih2013playing}, instead of a single approximation of state-action value functions, it leverages a minimum between $\eta^{\sl \text{\maxmin}}$ approximations.
That leads to the following temporal difference target
\begin{equation}
    r(s, a) + \gamma \max_{a'} \min_{i \in \{1 \ldots \eta^{\sl \text{\maxmin}}\}} \hat{Q}_{\psi_i} (s', a').
\end{equation}
Similarly to aforementioned hyper-parameters, the number of approximations $\eta^{\sl \text{\maxmin}}$ also requires per-environment hyper-parameter tuning.

While keeping the core approach we slightly modify the original implementation to isolate the effect of overestimation control from the impact of implementation details.
We explain how exactly and why in Appendix~\ref{app:mmql_mod}.

\section{Automating Bias Control}

\begin{table}[t!]
\resizebox{0.48\textwidth}{!}{%
\begin{tabular}{lll}
\toprule
\bf   \multirow{2}{*}{Base method} &\bf   Overestimation control&\bf Adaptive\\
& \bf technique & \bf method\\
\midrule
SAC (\citeauthor{haarnoja2018soft}) & \multirow{2}{*}{TQC (\citeauthor{kuznetsov2020controlling})} & \multirow{2}{*}{AdaTQC} \\
QR-DQN  (\citeauthor{dabney2018distributional})&&\\
\midrule
TD3 (\citeauthor{fujimoto2018addressing}) & WD3 (\citeauthor{he2020wd3}) & AdaWD3 \\
\midrule
DQN (\citeauthor{mnih2013playing}) &  \maxmin~(\citeauthor{lan2020maxmin}) &  Ada\maxmin\\
\bottomrule
\end{tabular}
}
\caption{The adaptive method proposed in this work is applied to truncated quantile critics (TQC), weighted delayed deep deterministic policy gradient (WD3), and Maxmin Q-learning (MMQL). The adaptive versions of these methods are named accordingly with Ada-prefix.}
\label{tab:methods}
\end{table}

\label{ABC} 
The mechanisms for controlling overestimation bias---as we show in the background section---often involves a single scalar hyper-parameter $\eta$, that controls the degree of overestimation reduction.
Tuning of $\eta$ is usually done with hyper-parameter optimization (e.g., grid search) and thus it requires multiple runs of an algorithm per environment, effectively reducing sample efficiency of the algorithm.

In the following section, we show that hyperparameter $\eta$ can be chosen with a single run without an outer hyperparameter optimization. 
We focus on the case when $\eta$ affects the overestimation bias monotonically~(e.g., increase of $\eta$ leads to a decrease of the bias).
That, combined with a data-driven estimation of an aggregated bias, allows us to adjust the over- and underestimation bias automatically.  
This automated control relies only on the monotonic dependence of the degree of overestimation compensation from control hyperparameter and is not tied to a specific learning technique~(e.g., TQC, WD3, \maxmin). 
We show how the resulting methods are built in Table~\ref{tab:methods}.


\definecolor{rd}{rgb}{0.94, 0.42, 0.32}
\definecolor{bl}{rgb}{0.32, 0.42, 0.94}
\subsection{How to Estimate Aggregated Bias}
\label{subsec:bias_est}
Estimation of the aggregated bias requires two components $i)$ a source of on-policy trajectories and $ii)$ an estimate of an unbiased state-action value function. 

As a source of \textit{near on-policy} trajectories, we use a small replay buffer $\mathcal{D}_\text{fresh}$. 
It contains trajectories from several recent policies and biases estimate toward previous policies.
To obtain an unbiased estimate of a value function we use $k$-step discounted sum of future rewards
\begin{equation}
\tilde{Q}(\rollout) 
= 
\sum_{i=t}^{\min(t+k-1, T-1)} \gamma^{i-t} r_i
+ 
\underbrace{\gamma^k \hat{Q}_\psi(s_{t+k}, a_{t+k})}_{\text{\bf{if}~}t+k<T},
\label{eq:est}
\end{equation}
where $\rollout=(s_t, a_t, r_t, \ldots, r_{\min(t+k-1, T-1)}, s_{\min(t+k, T)})$ is a rollout starting at a state $s_t$ and an action $a_t$. The rollout $\rollout$ contains at most $k$ future steps. $T$ is a length of the the episode representing \textit{done} signal. 

Combining them together we end up with the following estimate of the aggregated bias
\begin{equation}
    \wavyaggbias(\hat{Q}_\psi, \mathcal{B}) = \frac{1}{|\mathcal{B}|} \sum_{\rollout \in \mathcal{B}}   \Big(\hat{Q}_\psi(s_t, a_t) - \tilde{Q}(\rollout)\Big),
    \label{eq:be}
\end{equation}
where $\mathcal{B}$ is a mini-batch of rollouts sampled from $\mathcal{D}_\text{fresh}$ and $s_t, a_t$ are the starting state and action from rollout $\rho$.

If $\mathcal{D}_\text{fresh}$ contained only samples from the current policy and all episodes were played till the end ($k=\infty$)---which does not hold on practice---the estimate would be unbiased 
$$
    \aggbias(\hat{Q}_\psi, \pi) = 
    \underset{
    \substack{
        \mathcal{B} \sim \mathcal{D}_\text{fresh}\\
        \mathcal{D}_\text{fresh} \sim d_\pi(s, a)
    }
    }{\mathbb{E}}
    \wavyaggbias(\hat{Q}_\psi, \mathcal{B}).
$$
\begin{figure*}[t!]
\centering
\begin{subfigure}{0.33\textwidth}
\includegraphics[width=\textwidth]{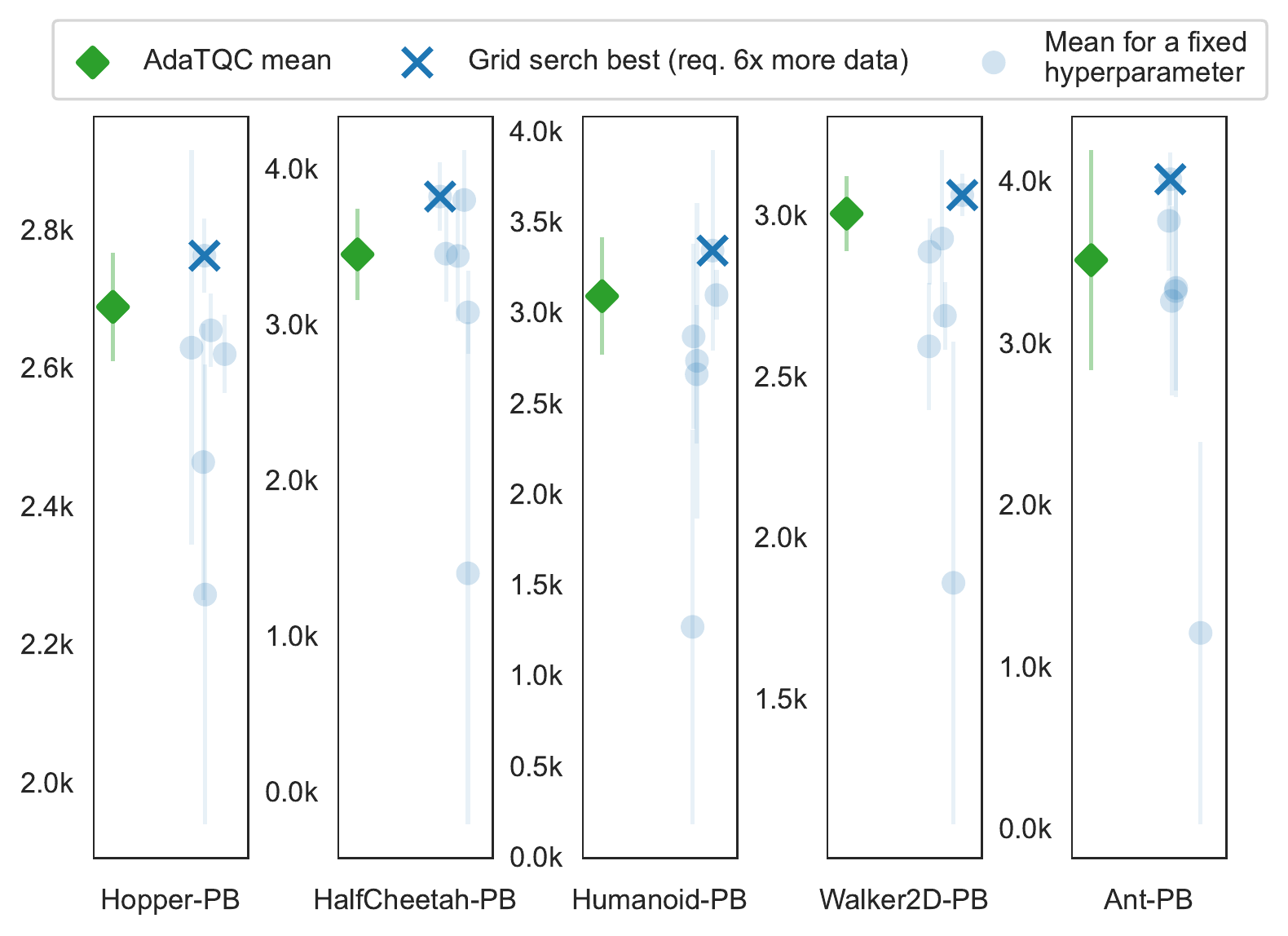}
\caption{AdaTQC (PyBullet)}
\label{fig:main_tqc_pybullet}
\end{subfigure}
\begin{subfigure}{0.33\textwidth}
\includegraphics[width=\textwidth]{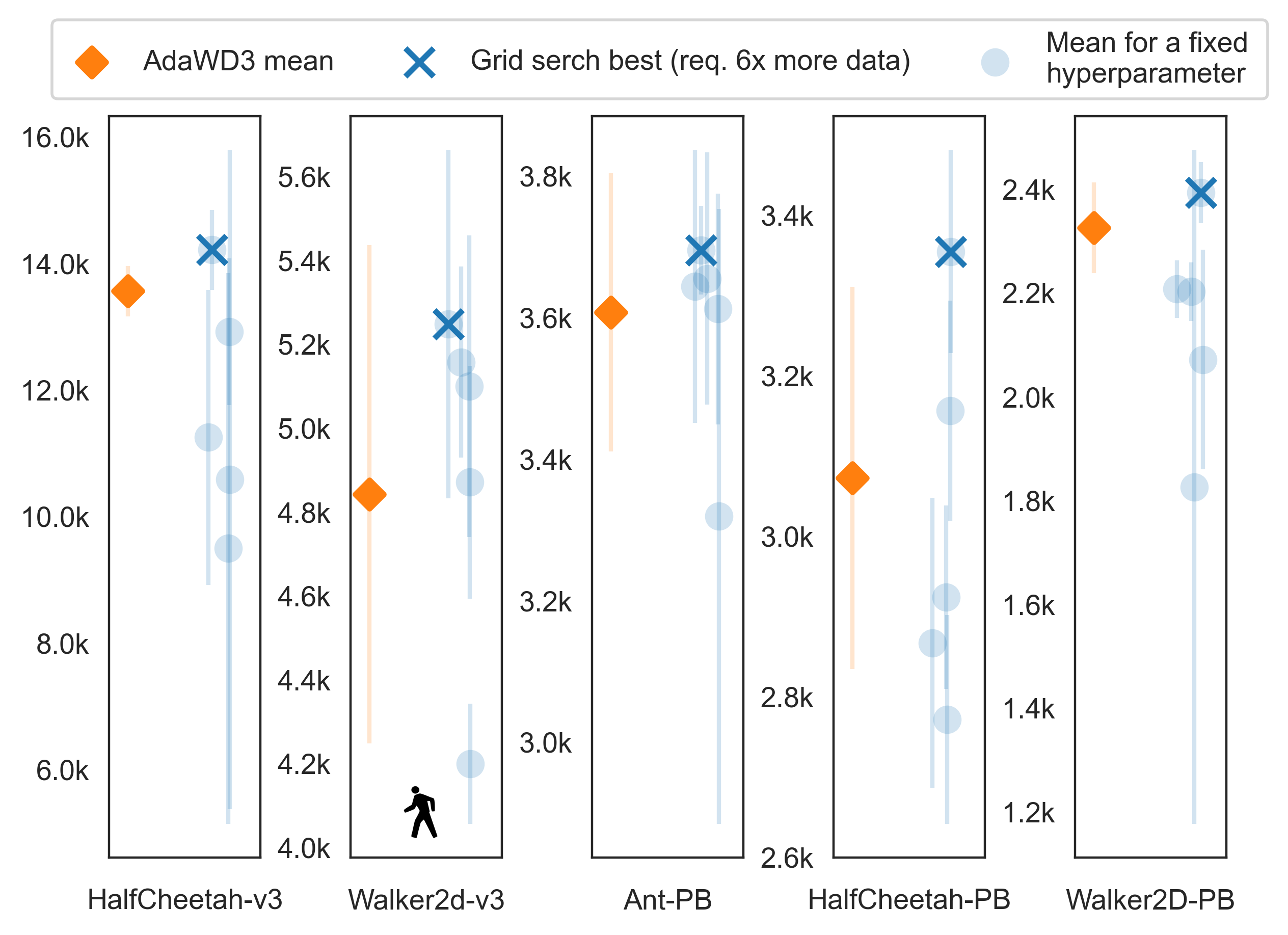}
\caption{AdaWD3}
\label{fig:main_adawd3}
\end{subfigure}
\begin{subfigure}{0.33\textwidth}
\includegraphics[width=\textwidth]{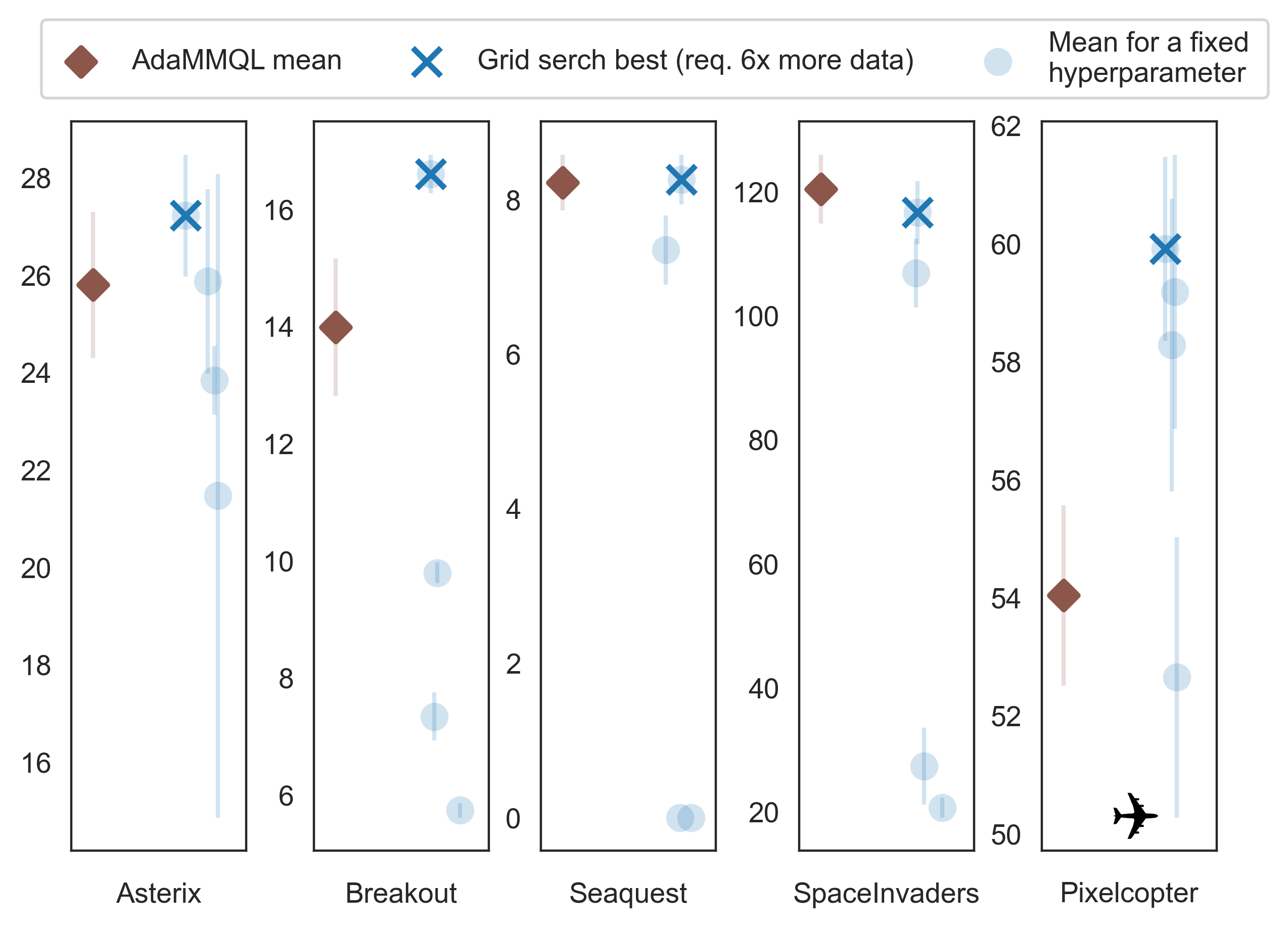}
\caption{MMQL}
\label{fig:main_mmql}
\end{subfigure}
\caption{Performance of AdaTQC on PyBullet (\subref{fig:main_tqc_pybullet}), AdaWD3 (\subref{fig:main_adawd3}) and \maxmin (\subref{fig:main_mmql}). 
Each line represents standard deviation across seeds.}
\label{fig:main_wd3}
\end{figure*}

\textbf{Environments with time limit.} More precisely, instead of $\mathcal{D}_\text{fresh}$ we use  $\mathcal{D}^{\text{valid}}_\text{fresh}$, that makes sure that there are enough future reward available. 
For example, in MuJoCo future rewards are not available for some state-action pairs, e.g., trajectories interrupted by time limit.
Putting away states with not enough future rewards  ($\leq k$) from $\mathcal{D}_\text{fresh}$ we obtain $\mathcal{D}^{\text{valid}}_\text{fresh}$. 
We require at least $k$ future rewards available. Rewards after \textit{done} are considered available and are equal to zero.

In the end, equation \ref{eq:be} uses the following approximations:
\begin{itemize}[noitemsep,topsep=0pt]
    \item Instead of states from $d_{\pi}(s)$ we use states from a replay buffer  $\mathcal{D}_\text{fresh}$, which contains data from recent policies.
    \item We use only states from $\mathcal{D}_\text{fresh}$ with enough  ($\geq k$) future rewards available which constitute $\mathcal{D}_\text{fresh}^{\text{valid}} \subseteq \mathcal{D}_\text{fresh}$.
    \item Return is estimated with $k$-step bootstrapped rewards.
\end{itemize}

\subsection{How to Adjust \texorpdfstring{$\eta$}{eta} During Training}

\begin{algorithm}[t!]
  \caption{\textit{Automatic bias control with  discrete $\eta$}\newline\label{alg:discrete}The algorithm requires following parameters:  
    fresh replay batch size $S_\text{fresh}$,
    fresh replay size in trajectories $N_\text{fresh}$, bias compute interval $M_\text{compute}$, $\eta$ update interval $M_\text{update}$.
    bias smoothing parameter $\gamma_\eta$. Function \texttt{Update} is an update step of any suitable RL algorithm, $\pi$ may be implicit as in MMQL.}
\begin{algorithmic}
    \STATE Initialize experience replay 
    \STATE $\mathcal{D}_{\text{fresh}} = \text{Replay}(\text{max\_size} = N_\text{fresh})$
    \STATE Initialize $\eta$, policy $\pi_\phi$, critic $\hat{Q}_\psi$, smoothed bias $B_\text{smooth} = 0$
    \FOR{each iteration $t$}
        \STATE collect transition $(s_t, a_t, r_t, s_{t+1})$ with $\pi_\phi$
        \STATE $\mathcal{D}_\text{fresh} \leftarrow \mathcal{D}_\text{fresh} \cup \{ (s_t, a_t, r_t, s_{t+1}) \}$
        \STATE $\pi_\phi, \hat{Q}_\psi \leftarrow \text{Update} (\pi_\phi, \hat{Q}_\psi, \eta)$
        \STATE \textbf{once per} $M_\text{compute}$ \textbf{iterations}
            \STATE \quad $\mathcal{B} \leftarrow$ sample a batch  of $S_\text{fresh}$ \textit{valid} rollouts $\rho$ 
            \STATE \quad \quad from  $\mathcal{D}_\text{fresh}$
            \STATE \quad $B_\text{smooth} = \gamma_\eta B_\text{smooth} + (1 - \gamma_\eta) \wavyaggbias(\hat{Q}_\psi, \mathcal{B})$
        \STATE \textbf{once per}
        $M_\text{update}$  \textbf{iterations}
            \STATE \quad $\eta = \eta + \sign (B_\text{smooth})$
    \ENDFOR
    \RETURN policy $\pi_\phi$, critic $\hat{Q}_\psi$
\end{algorithmic}
\end{algorithm}

We denote $\eta_t(\pi_t, \hat{Q}_t$) at optimization step $t$ as $\eta_t$. 
We assume that the bias at step $t+1$ depends on $\eta_t$ through the parameters $\psi_{t+1}$ of~$\hat{Q}_{t+1}$.
Specifically, increase of $\eta_{t}$ reduces the overestimation at step $t+1$, while decrease of $\eta_{t}$ relaxes the intensity of correction at step $t+1$. 

To automate bias control we introduce \textit{bias control updates}:
\begin{minipage}[b]{0.23\textwidth}
    \begin{gather*}
        \textbf{if }\aggbias(\hat{Q}_t, \pi_t) > 0 \\[-3pt]
        \text{\textcolor{rd}{overestimation}}\\[-3pt]
        \downarrow\\[-3pt]
        \eta_{t+1}=\eta_{t}+\lambda \\[-3pt]
        \text{increase }\eta
        \\[-3pt]
        \downarrow\\[-3pt]
        \text{less overestimation}\\[-3pt]
        \text{at step }t+1
    \end{gather*}
\end{minipage}
\begin{minipage}[b]{0.23\textwidth}
    \begin{gather*}
        \textbf{if }\aggbias(\hat{Q}_t, \pi_t) < 0 \\[-3pt]
        \text{\textcolor{bl}{underestimation}}\\[-3pt]
        \downarrow\\[-3pt]
        \eta_{t+1} = \eta_{t}-\lambda\\[-3pt]
        \text{decrease }\eta
        \\[-3pt]
        \downarrow\\[-3pt]
        \text{less underestimation}\\[-3pt]
        \text{at step }t+1
    \end{gather*}
\end{minipage}

where $\lambda$ is a step size. Note that algorithm can work with both continuous and discrete $\eta$, with $\lambda=1$ in the latter case. In practice we calculate $\wavyaggbias$ once every $M_\text{compute}$ environment steps to reduce computational costs. Algorithm \ref{alg:discrete} shows discrete version of automatic bias control, while continous one can be found in Appendix \ref{sec:continuous_alg}.

For continuous hyperparameters the \textit{bias control updates} can be understood as the approximation of the following meta-gradient step 
\begin{equation} \label{eq:eta-update}
    \eta_{t+1} = \eta_t +\lambda \!\sign\!\left(\!\nabla_\eta \aggbias(\hat{Q}_{\psi_{t+1}(\eta)}, \pi) |_{\eta=\eta_t}\!\right),
\end{equation}
where $\sign(\cdot)$ serves as the normalization, that allows to select step-size $\lambda$ independently of the absolute value of returns.
The approximation is possible due to the prerequisite on the monotonic influence of $\eta$ on bias that effectively supplies information on the sign of the gradient.

The procedure provides a way for data-driven control of the overestimation bias.
In contrast to the pre-defined policy, the correction can adapt to the current level of overestimation during the training.
\begin{figure*}[t!]
\centering
\includegraphics[width=\textwidth]{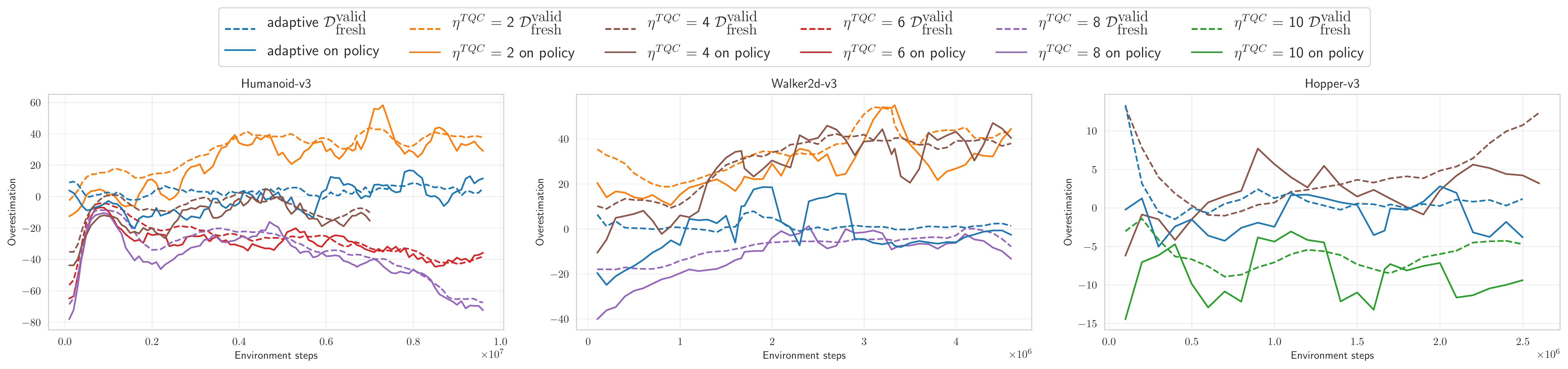}
\caption{
Unbiased estimate of the aggregated bias using additional data (on policy) and biased estimate based on $\mathcal{D}^{\text{valid}}_{\text{fresh}}$ for TQC and AdaTQC.
 $\mathcal{D}^{\text{valid}}_{\text{fresh}}$-based estimate of the aggregated bias closely matches the unbiased on-policy version of the estimate. With adaptive control of $\eta^{TQC}$ the bias remains close to zero, confirming that AdaTQC is indeed able to control the bias. Data points are calculated each 100k steps, the curves are averaged over 4 runs and smoothed with window of 5. For visual clarity we omit some values, plot with full $\eta^{TQC}$ grid is in Appendix \ref{sec:bias_eval_app}.}
\label{fig:estimate}
\end{figure*}

\begin{figure*}[t!]
\centering
\includegraphics[width=\textwidth]{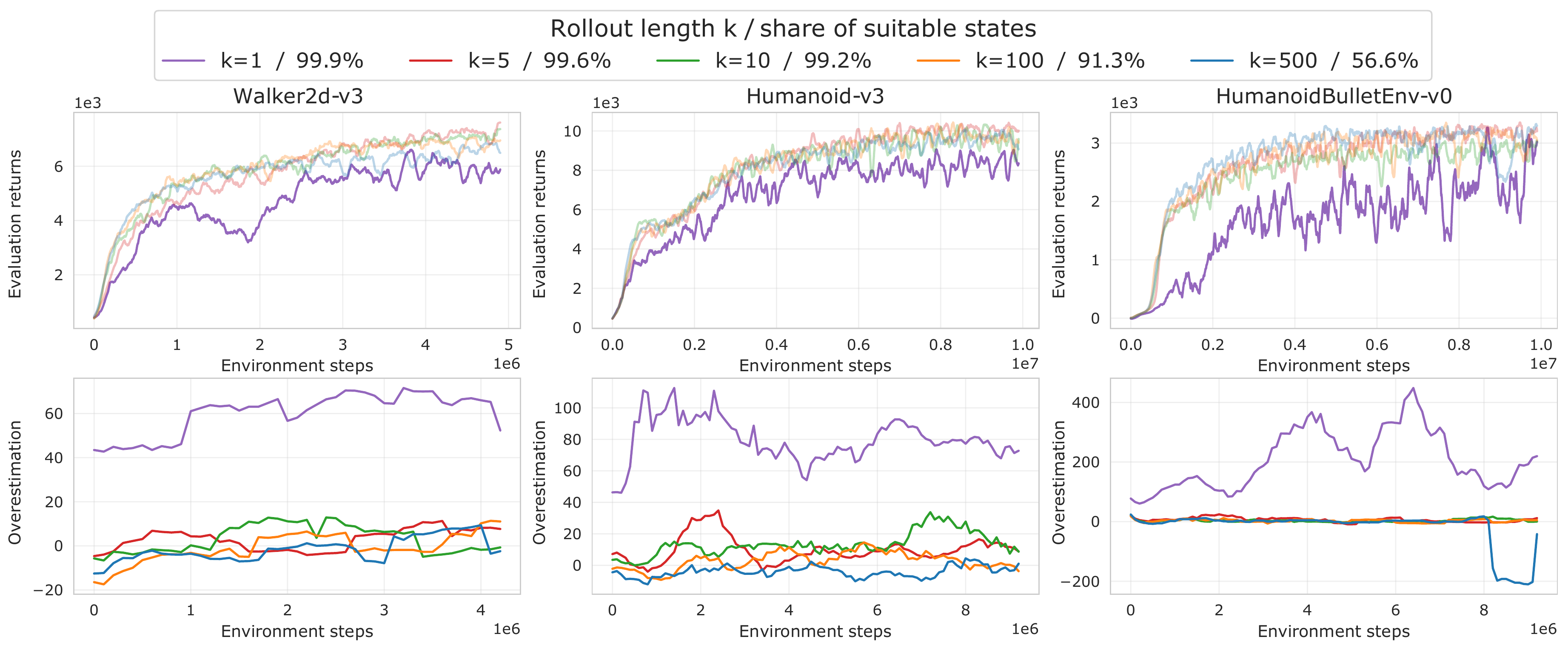}
\caption{
Performance, aggregated bias calculated on-policy and share of suitable states $|\mathcal{D}_\text{fresh}^\text{valid}|/|\mathcal{D}_\text{fresh}|$ for different values of 
rollout length $k$.
}
\label{fig:rollout_length}
\end{figure*}
\section{Experiments}
We apply the proposed automatic bias correction method over three overestimation control algorithms: TQC, WD3 and \maxmin.
For TQC and WD3 we run experiment on MuJoCo~\citep{todorov2012mujoco} and on PyBullet~\citep{coumans2021}, for \maxmin~on MinAtar~\cite{Young2019MinAtarAA} and  PyGame Learning Envs (PLE)~\cite{tasfi2016PLE}. 

In this section we aim to demonstrate the following:
\begin{enumerate}[noitemsep,topsep=0pt]
    \item[$a)$] The proposed automatic bias control is competitive with an otherwise \textit{essential} grid search, while require significantly less interactions. 
    \item[$b)$] The $\mathcal{D}^{\text{valid}}_{\text{fresh}}$-based estimate of an overestimation bias closely resembles the true on-policy bias. We also make a connection between the number of future rewards $k$ and the quality of the bias correction.
    \item[$c)$] In the result, the proposed method successfully optimizes an overestimation bias during a single run.
    \item[$d)$] $\eta$ behaves reasonably during training and often concentrates near optimal values obtained by grid search.
\end{enumerate}

All models are implemented\footnote{We publish source code anonymously at \href{https://github.com/icml2022-submission/ada_tqc}{github repository 1} for AdaTQC and at \href{https://github.com/icml2022-submission/ada_maxmin}{github repository 2} for Ada\maxmin} in PyTorch~\citep{NEURIPS2019_9015}. 
To evaluate the performance of an algorithm we use an average over the last $10^5$ steps.
We list implementation details and hyperparameters in Appendix~\ref{app:hyp}. 
Full optimization curves for AdaTQC, AdaWD3 and AdaMMQL are located in Appendix \ref{sec:learning_curves}.

\subsection{Performance comparison}
\label{subsec:perf_comp}
We compare the performance of the proposed automatic bias control technique against the grid search. 
We keep the grid close to the one used in the original works.

\textbf{AdaTQC~} We adjust the number of the dropped quantiles $\eta^{TQC}$ in AdaTQC, and use $\eta^{TQC} \in \{0,2,4,6,8,10\}$ as a grid for TQC. For the number of critics we use $N=2$.
The results are in Figure~\ref{fig:ada_tqc} for MuJoCo and in Figure~\ref{fig:main_tqc_pybullet} for PyBullet.
Performance of AdaTQC is always better than the average over hyperparameters and similar to the second to best performing hyperparameter which is different for different environments.

\textbf{AdaWD3~} Grid for the weight of the minimum in WD3 is $\eta^{\text{WD3}}\in \{0, 0.25, 0.5, 0.75, 1\}$.
Performance for MuJoCo and PyBullet is presented in Figure~\ref{fig:main_adawd3}.
AdaWD3 is competitive with the only exception of MuJoCo Walker environment~(\walkersym).

\begin{figure}[t!]
\centering
\begin{subfigure}{0.4\textwidth}
\includegraphics[width=\textwidth]{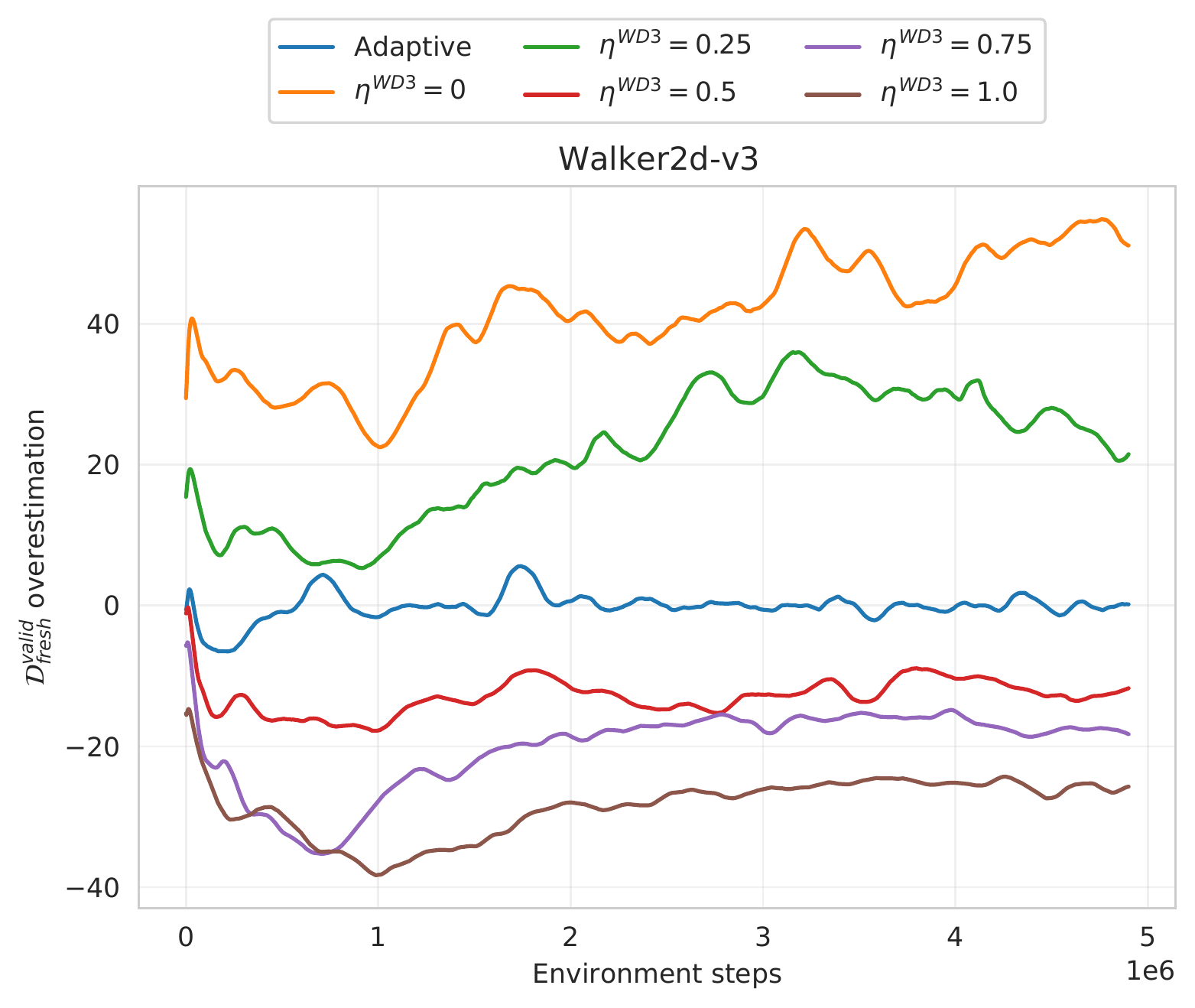}
\caption{For WD3 on \textit{Walker2d-v3}}
\label{fig:bias_wd3}
\end{subfigure}
\begin{subfigure}{0.4\textwidth}
\includegraphics[width=\textwidth]{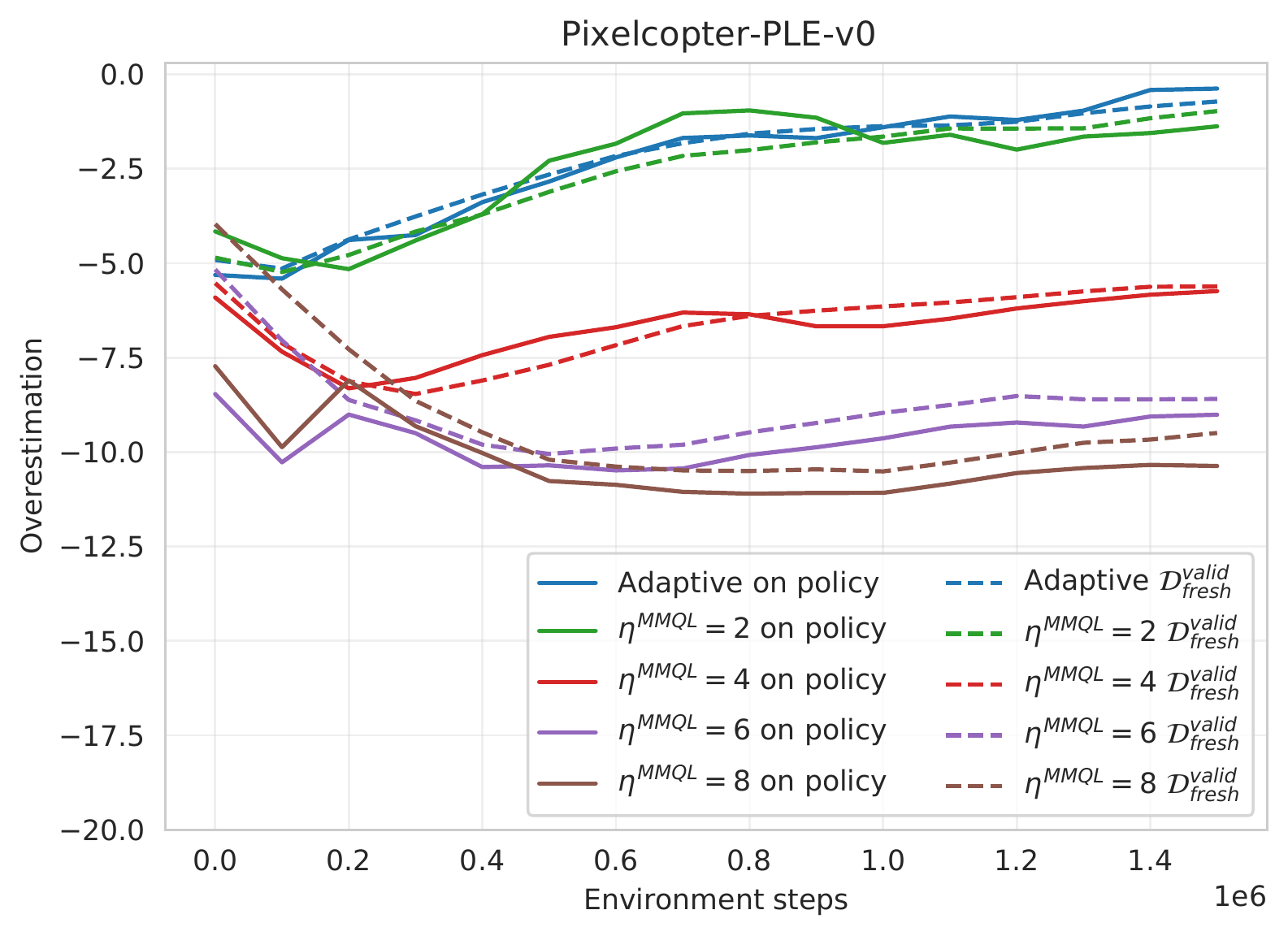}
\caption{For \maxmin~on \textit{Pixelcopter}}
\label{fig:bias_maxmin}
\end{subfigure}
\caption{Bias estimates.}
\end{figure}

\begin{figure*}[t!]
\centering
\begin{subfigure}{\textwidth}
\includegraphics[width=\textwidth]{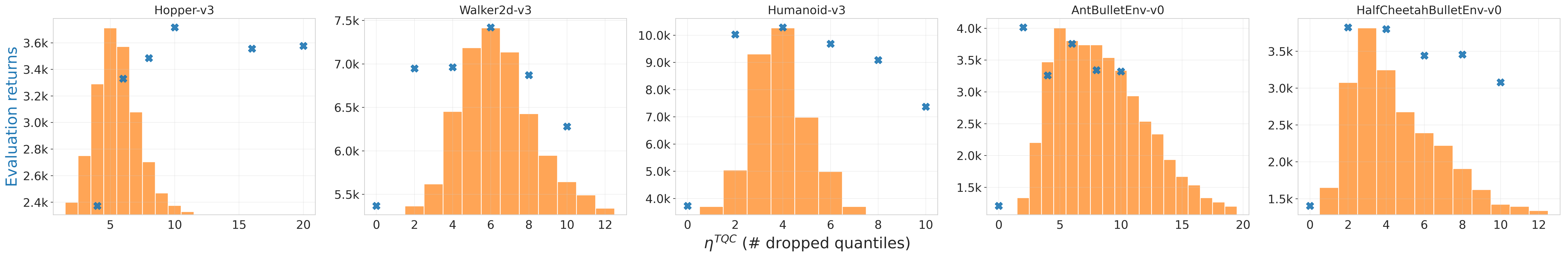}
\caption{A \textcolor{orange}{\bf histogram} of values of $\eta^{TQC}$ for AdaTQC, and values of performance for fixed hyperparameters for non-adaptive TQC (\textcolor{bl}{\xmark})
}
\label{fig:eta_tqc}
\end{subfigure}
\begin{subfigure}{\textwidth}
\includegraphics[width=\textwidth]{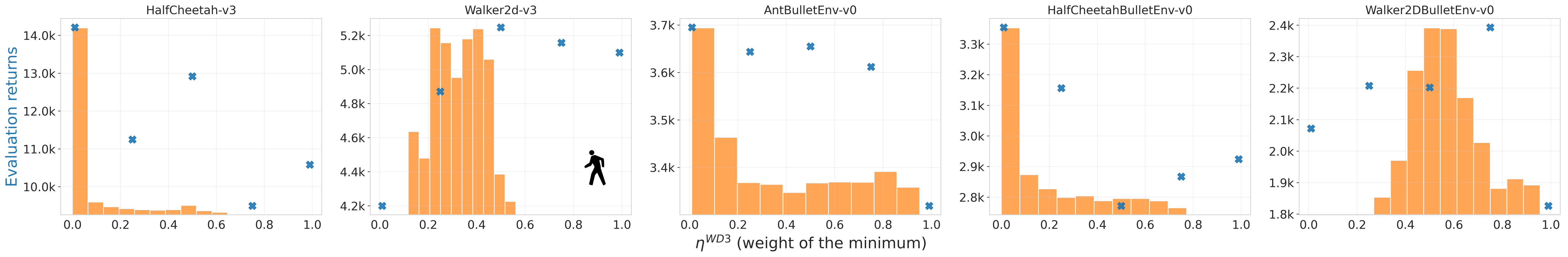}
\caption{A \textcolor{orange}{\bf histogram} of values of $\eta^{WD3}$ for AdaWD3, and values of performance for fixed hyperparameters for non-adaptive WD3 (\textcolor{bl}{\xmark})}
\label{fig:eta_wd3}
\end{subfigure}
\begin{subfigure}{\textwidth}
\includegraphics[width=\textwidth]{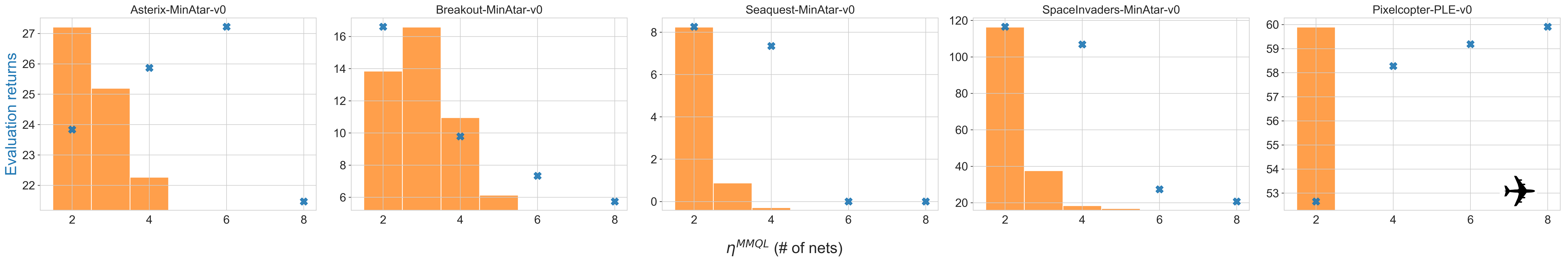}
\caption{A \textcolor{orange}{\bf histogram} of values of $\eta^{MMQL}$ for AdaMMQL, and values of performance for fixed hyperparameters for non-adaptive MMQL (\textcolor{bl}{\xmark})}
\label{fig:eta_maxmin}
\end{subfigure}
\caption{\textcolor{orange}{\bf Histograms} of values of hyperparameter $\eta$ collected throughout a training of adaptive methods, and the \textcolor{bl}{\bf performances} of non-adaptive method with a fixed hyperparameter $\eta$. In the majority of cases the pick of the histogram lays in the are of \textit{good} values of controlled hyperparameters, that correspond to a sophisticated final performance. }
\label{fig:eta_hist}
\end{figure*}

\walkersym~~Interestingly, performance of AdaWD3 on Walker2d is considerably worse than the performance of fixed hyperparameters $\eta^{WD3} \in \{0.5, 0.75, 1\}$ (Figure~\ref{fig:main_adawd3}). 
At the same time, AdaWD3 successfully optimizes bias unlike the aforementioned fixed hyperparameters (Figure~\ref{fig:bias_wd3}) while moving control hyperparameter to the worse performing value  (Figure~\ref{fig:eta_hist}). 
We attribute this fact to the fundamental limitation of our method---it performs sub-optimally if performance of underlying bias correction technique is not tied to the low absolute bias.
tSpecifically, all the underlying methods uses homogeneous (i.e. the same) control policy across all states, which could be not flexible enough.

\textbf{Ada\maxmin~}  Grid for the number of Q-networks inside the minimum for \maxmin~is $\eta^{\maxmin} \in \{2, 4, 6, 8\}$. 
Performance for MinAtar and PLE (Pixelcopter) is presented in Figure~\ref{fig:main_mmql}.
Ada\maxmin~is competitive with the only exception of Pixelcopter environment (\coptersym).

\coptersym~We see the same picture as for AdaWD3/Walker (\walkersym) -  bias is optimized better (Figure~\ref{fig:bias_maxmin}), but solutions with bigger absolute bias ($\eta^{\maxmin} \in \{4, 6, 8\}$) perform better (Figure~\ref{fig:eta_maxmin}).
\subsection{Sample Efficiency}
In this section, we quantify the benefits from the absence of hyperparameter tuning. Specifically, we answer the following question: {\sl How many tries of grid search do we need to achieve the performance of the adaptive method?}

We average improvement in sample efficiency and show the results in Table \ref{table:sample_eff_tqc}. 
A value $x$ means that non-adaptive method on average needs at least $x$ full runs with different hyperparameters to achieve the same level of performance as adaptive method with one run.
For example $\times3$~means that one needs to try at least $3$ values of hypereparameter for the non-adaptive method and maximize over them to achieve the same performance as adaptive method does with just a single run.
The details on sample efficiency computation are presented in Appendix~\ref{app:sample_eff_comp}.

\begin{table}[t!]
\begin{subtable}[t]{0.23\textwidth}
\resizebox{1.\textwidth}{!}{
\begin{tabular}[t]{lc}
\toprule
Environment& ISE ($\times$)~\textuparrow\\
\midrule
Asterix      & $2$               \\
Breakout         & $3$ \\
Seaquest    & $4$               \\
SpaceInvader    & $>4$               \\
Pixelcopter \coptersym & $1$               \\
\bottomrule
\end{tabular}
}\caption{Ada\maxmin}
\end{subtable}
\begin{subtable}[t]{0.23\textwidth}
\vspace{1cm}
\resizebox{1.\textwidth}{!}{
\begin{tabular}[t]{lc}
\toprule
Environment& ISE ($\times$)~\textuparrow\\
\midrule
HalfCheetah      & $\textcolor{white}{\geq}2$               \\
Hopper      & $\textcolor{white}{\geq}3$               \\
Ant         & $>6$ \\
Humanoid    & $\textcolor{white}{\geq}4$               \\
Walker2d    & $\textcolor{white}{\geq}2$               \\
HalfCheetah-B & $\textcolor{white}{\geq}2$               \\
Hopper-B              & $\textcolor{white}{\geq}3$ \\
Ant-B                  & $\textcolor{white}{\geq}2$               \\
Walker2d-B             & $\textcolor{white}{\geq}4$               \\
Humanoid-B             & $\textcolor{white}{\geq}3$            \\
\bottomrule
\end{tabular}
}\caption{AdaTQC}
\end{subtable}
\begin{subtable}[t]{0.23\textwidth}
\vspace{-2.5cm}
\resizebox{1.\textwidth}{!}{
\begin{tabular}[t]{lc}
\toprule
Environment& ISE ($\times$)~\textuparrow\\
\midrule
Ant      & $2$               \\
Walker2d \walkersym        & $1$ \\
HalfCheetah    & $4$               \\
HalfCheetah-B    & $2$               \\
Walker2d-B & $4$               \\
\bottomrule
\end{tabular}
}\caption{AdaWD3}
\end{subtable}

\caption{
Average improvement in sample efficiency (ISE ~\textuparrow higher is better) of~\method in comparison to TQC. 
All the runs consist of the same amount of interactions.
} \label{table:sample_eff_tqc}
\end{table}

\subsection{Process of \texorpdfstring{$\eta$}{eta} adaptation}
On Figure~\ref{fig:eta_hist} we show the distribution of values of control parameter $\eta$ during adaptive training for AdaTQC, AdaWD3 and Ada\maxmin.
In general, adaptive $\eta$ gravitates towards best performing constant $\eta$-s on most of the environments.
However, there are cases like AdaWD3/Walker~(\walkersym) and Ada\maxmin/Pixelcopter~(\coptersym) which do not follow this pattern.
We discuss possible reasons in Section~\ref{subsec:perf_comp}.

\subsection{Quality of Bias Estimate}
\label{subsec:vis_bias}
In the end of Section~\ref{subsec:bias_est} we describe several approximations used in the bias estimate. 
To show the quality of the proposed estimate we compare it against the unbiased estimate that is based on on-policy data.
In Figure~\ref{fig:estimate} we plot them side-by-side for a subset of MuJoCo environments.
Despite the approximations, the proposed estimate closely follows the unbiased one.

\subsection{Number of future steps in return estimate}
\label{subsec:n_steps}

In environments conventionally modified with time limit such as MuJoCo, PyBullet and PLE Pixelcopter all the future rewards usually are not available since good policy tends to reach the aforementioned time limit.
As we explained in Section~\ref{subsec:bias_est} it leads to the need of selection of number of steps $k < \infty$ in return estimate (equation~\ref{eq:est}).

The hyperparameter $k$ influences the algorithm as follows $i)$ small values of $k$ could make the return estimate severely biased while $ii)$ big values of $k$ could significantly decrease the number of available states biasing state visitation distribution in $\mathcal{D}_\text{fresh}^\text{valid}$.
How significant are these effects in practice?

To answer this question we quantify these processes. For different $k$ we  compute a share of suitable states as the fraction $S=|\mathcal{D}_\text{fresh}^\text{valid}|/|\mathcal{D}_\text{fresh}|$ that is nearly constant throughout the training. $S$ shows the percent of states where at least $k$ future rewards are available. Also we plot the unbiased estimate of overestimation bias. 
The results are available in Figure~\ref{fig:rollout_length}.

We can see that only $k=1$ fails to control the bias, moreover it has significant performance degradation, while the greater values perform similarly. 
So ($i$) extremely small value of $k$  is likely to considerably bias the estimate, while ($ii$) the bias from reduction the number of states  does not significantly affect the system.

\section{Related work}
Accurate estimation of the value function is one of the core problems in reinforcement learning.
The use of approximation-based algorithms induces a consistent and uneven overestimation~\citep{thrun1993issues} which negatively affects their performance~\citep{thrun1993issues, szita2008many, strehl2009reinforcement, van2016deep, lan2020maxmin}.
This motivated the development of bias correction techniques.

A wide group of overestimation correction techniques is focused on the development of new estimates of the temporal-difference target. 
Double DQN~\citep{van2016deep} disentangles the action selection from the estimation of returns by using different Q-functions.
Averaged Q-learning~\citep{Anschel2017AveragedDQNVR} takes the average of previously learned state-action values to reduce the overestimation bias.
Maxmin Q-learning~\citep{lan2020maxmin} uses an ensemble of Q-functions---it varies the number of critic networks in the minimum---to balance between over-and underestimation.

The target modification is also widely applied in continuous control. 
Twin Delayed DDPG~\citep{fujimoto2018addressing} takes a minimum of two independent Q-functions.
Although successfully eliminating overestimation bias it may cause a severe underestimation bias~\citep{pan2020softmax, he2020reducing} which deteriorates the performance of the algorithm due to the pessimistic underexploration~\citep{Ciosek2019BetterEW}.
Triplet-average DDPG~\citep{wu2020reducing} and Weighted Delayed DDPG~\citep{he2020reducing} use a weighted sum of the minimum of Q-functions and the average of an ensemble of Q-functions to mitigate the effect of underestimation.

The other techniques for bias correction include: double actors~\citep{pan2020softmax, lyu2021efficient} and distributional~\citep{ma2020dsac, duan2021distributional} methods.
Truncated quantile critics (TQC)~\citep{kuznetsov2020controlling} has promoted this ides further and used a truncated ensemble of distributional critics for more fine grained control of the bias. 

The aforementioned methods control the bias via a pre-defined policy for which both strategy and magnitude of the correction are defined prior to training. Furthermore, such a policy requires tuning of the magnitude of the correction~\citep{kuznetsov2020controlling, lan2020maxmin, zhang2017weighted}.

In this work, we show that the overestimation bias can be controlled during training in an adaptive data-driven way. 
This results in a significant reduction of the actual number of required environment interactions when accounting for the otherwise essential hyper-parameter search.

There are several closely related papers also suggesting data-driven adaptation of control hyperparameter. Tactical optimism and pessimism~\citep{Moskovitz2021TacticalOA} employs a multi-armed bandit, which aims to maximize immediate performance improvements, to balance between optimistic and pessimistic strategies. In contrast, our method relies directly on the bias estimation to adjust control hyperparameter.
GPL-SAC~\citep{cetin2021learning} proposed a new overestimation control technique based on uncertainty estimates and adaptation technique simultaneously. Adaptation technique represents particular case ($k=1$) of our approach. 
Our experiments (Section~\ref{subsec:n_steps}) show that $k=1$ is considerably inferior in terms of bias estimation and results in a significant drop in performance.
Adaptively Calibrated Critic~\citep{dorka2021adaptively} is a concurrent method presenting essentially the same idea as this paper while using slightly different optimization procedure.

\section{Conclusion} 
This work introduces a method for on-the-fly correction of overestimation bias. 
It is suitable for a significant class of overestimation control techniques.
We applied this approach to TQC, WD3, and Maxmin Q-learning. 
Empirically, we show that the proposed approach can automatically reduce the bias in a single run while preserving the performance without a costly outer loop hyperparameter tuning.

\textbf{The limitations} of the proposed approach include:
\begin{itemize}[noitemsep,topsep=0pt]
    \item[$i)$] The fundamental limitation of the proposed approach is its inability to reliably estimate and correct unaggregated biases. Strictly speaking, a large aggregated bias does not necessary imply performance degradation, and the best possible option is to correct for state-action dependent biases. That, unfortunately, may be as hard as estimate Q-function.
    \item[$ii)$] The proposed adaptation technique is not able to achieve optimal results if underlying bias correction method performs better with non-zero aggregated bias~(\coptersym, \walkersym).
    \item[$iii)$] Ignoring last $k$ states in episodes that have reached time limit during bias estimate aggravates already presented in replay buffer bias towards the start of the episode. However, for wide range of $k$ it doesn't hurt neither performance, nor bias estimate.
\end{itemize}

We see state-dependent overestimation control parameters $\eta(s)$ as a promising direction for the future research.
It could allow to reduce bias in each state individually and the proposed approach is ready ``as is'' for such modification. 

\bibliography{example_paper}
\bibliographystyle{icml2022}

\clearpage
\onecolumn

\begin{appendices}

\section{Dependence on the value of control hyperparameter}
\label{app:heatmap}

\begin{figure*}[h!]
\centering
\begin{subfigure}{0.32\textwidth}
\includegraphics[width=\textwidth]{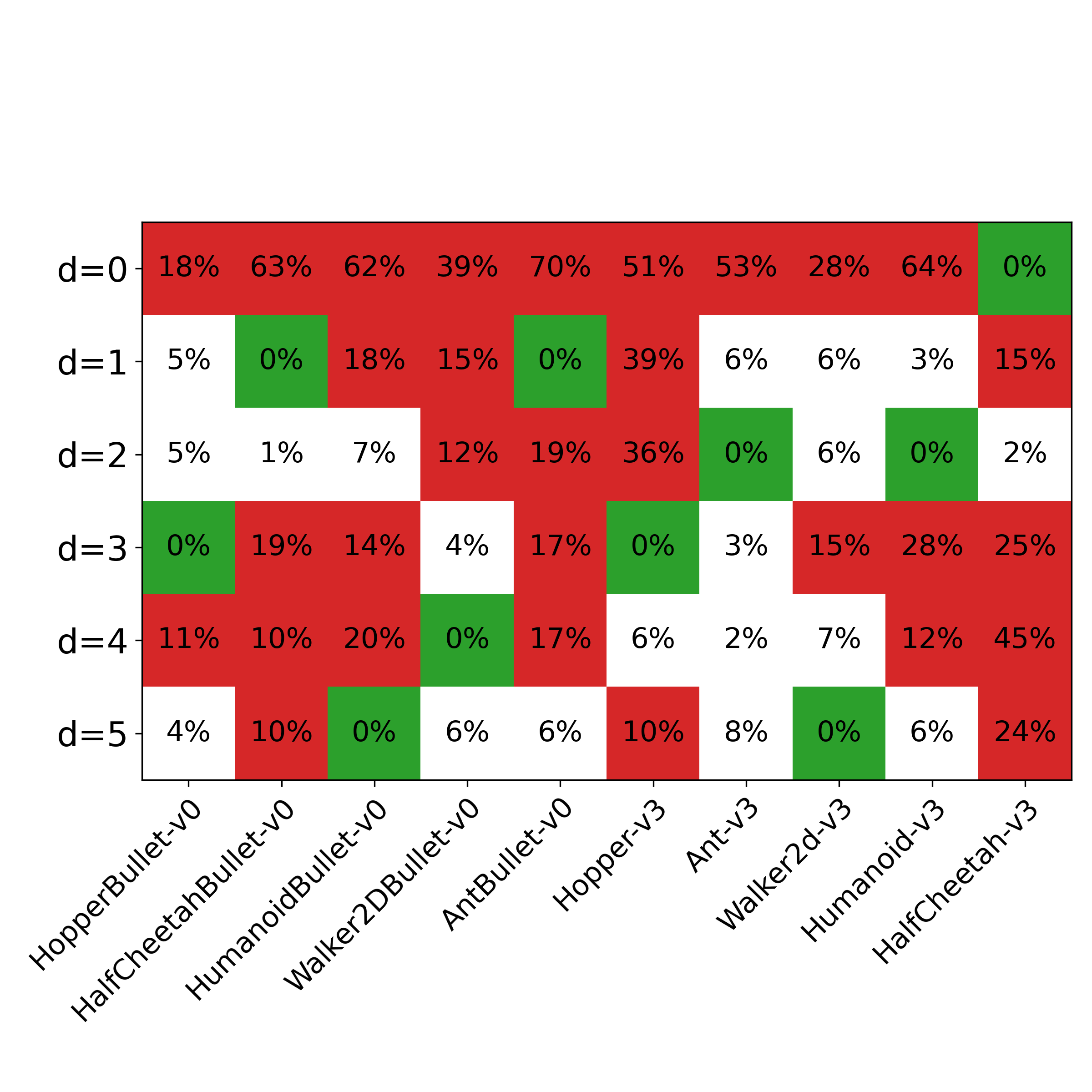}
\caption{TQC}
\label{fig:heatmap_tqc}
\end{subfigure}
\begin{subfigure}{0.32\textwidth}
\includegraphics[width=\textwidth]{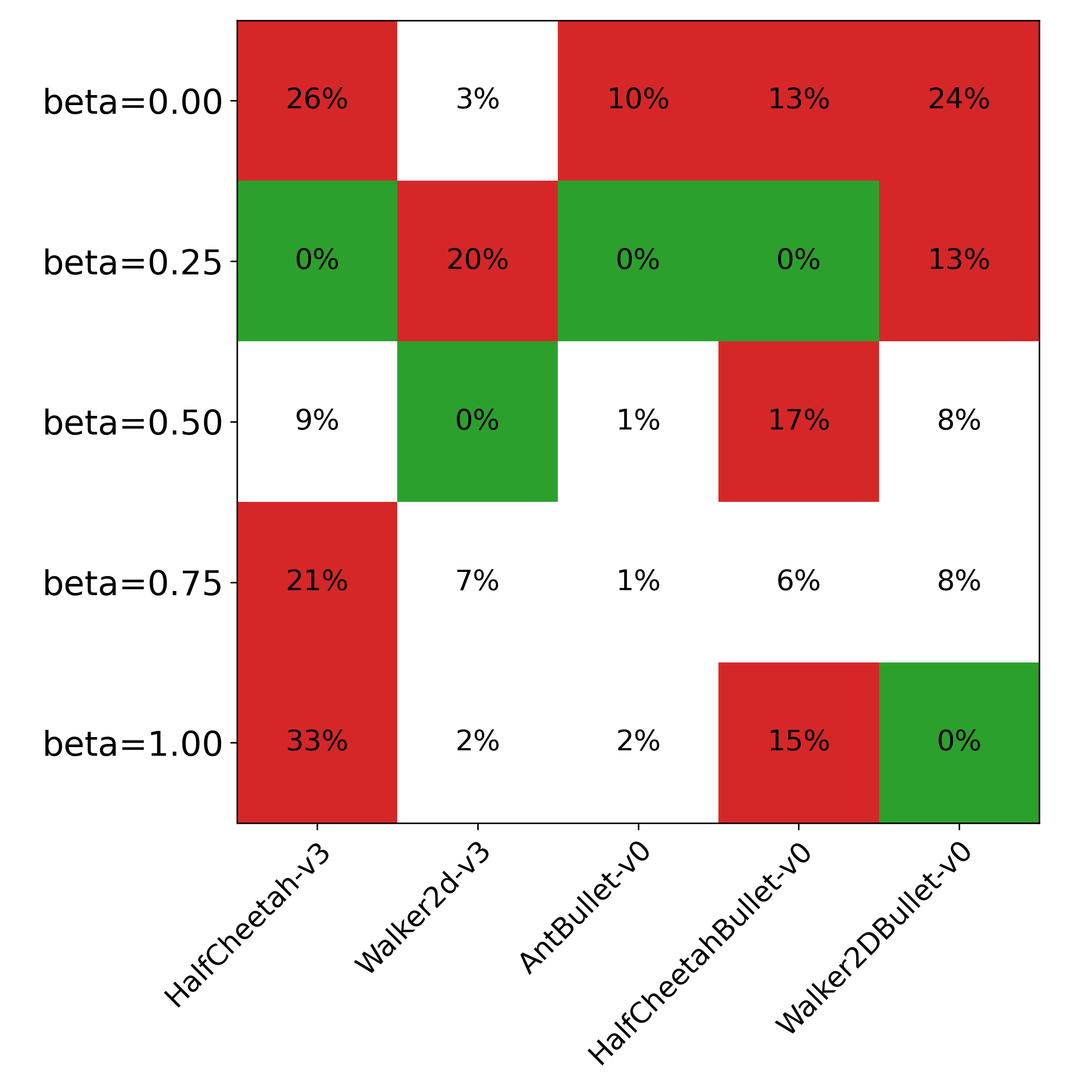}
\caption{WD3}
\label{fig:heatmap_wd3}
\end{subfigure}
\begin{subfigure}{0.32\textwidth}
\includegraphics[width=\textwidth]{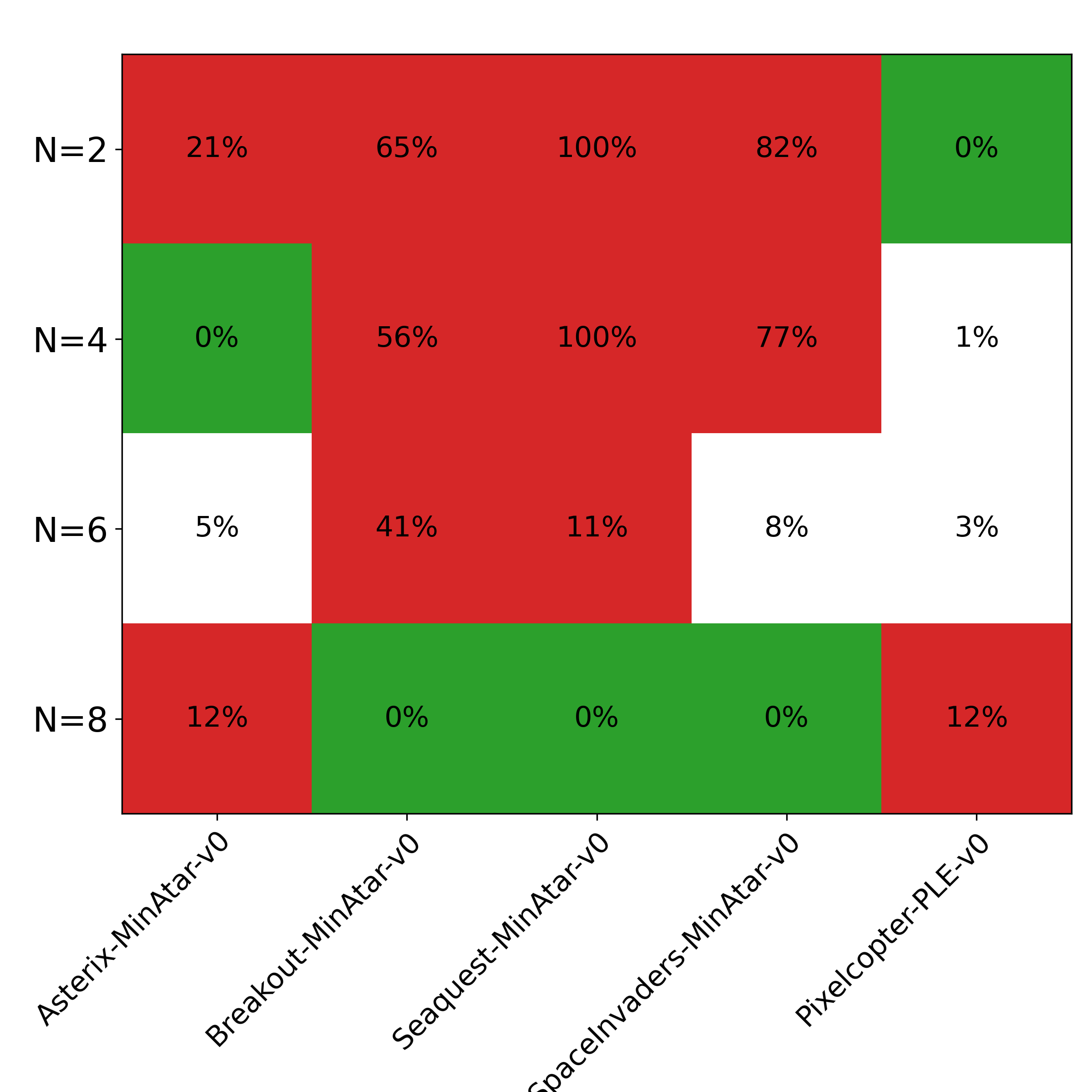}
\caption{\maxmin}
\label{fig:heatmap_maxmin}
\end{subfigure}
\caption{Relative performance decrease of corresponding method with different values of hyperparameter w.r.t. the performance of the best one for every environment. The best value of hyperparameter for every environment is denoted by green. The optimal value varies dramatically across environments, while there is no good universal choice of it.
}
\label{fig:heatmap}
\end{figure*}

\section{Adaptive bias control for the continuous control parameter \texorpdfstring{$\eta$}{eta}}
\label{sec:continuous_alg}

\begin{algorithm}[h]
  \caption{\textit{Automatic bias control with  continuous $\eta$ parameter}
  \label{alg:continuous}
  The algorithm requires following parameters:  
    fresh replay batch size $S_\text{fresh}$,
    fresh replay size in trajectories $N_\text{fresh}$, bias compute interval $M_\text{compute}$, $\eta$ learning rate $\lambda$.}
\begin{algorithmic}
    \STATE Initialize experience replay 
    \STATE $\mathcal{D}_{\text{fresh}} = \text{Replay}(\text{max\_size} = N_\text{fresh})$
    \STATE Initialize $\eta$, policy $\pi_\phi$, critic $\hat{Q}_\psi$, smoothed bias $B_\text{smooth} = 0$
    \FOR{each iteration $t$}
        \STATE collect transition $(s_t, a_t, r_t, s_{t+1})$ with $\pi_\phi$
        \STATE $\mathcal{D}_\text{fresh} \leftarrow \mathcal{D}_\text{fresh} \cup \{ (s_t, a_t, r_t, s_{t+1}) \}$
        \STATE $\pi_\phi, \hat{Q}_\psi \leftarrow \text{Update} (\pi_\phi, \hat{Q}_\psi, \eta)$
        \STATE \textbf{once per} $M_\text{compute}$ \textbf{iterations}
            \STATE \quad $\mathcal{B} \leftarrow$ sample a batch  of $S_\text{fresh}$ \textit{valid} rollouts $\rho$ from  $\mathcal{D}_\text{fresh}$
            \STATE \quad $\eta = \eta + \lambda \sign \wavyaggbias(\hat{Q}_\psi, \mathcal{B})$
    \ENDFOR
    \RETURN policy $\pi_\phi$, critic $\hat{Q}_\psi$
\end{algorithmic}
\end{algorithm}

\section{Experimental setup and hyperparameters}
\label{app:hyp}
In our experiments, we observed that the use of \texttt{MuJoCo 2.0} with versions of \texttt{Gym} at least up to \texttt{v0.15.4} nullifies state components that correspond to contact forces which makes the results incomparable with the previous work. All the experiments thus were done with \texttt{MuJoCo 1.5} and \texttt{v3} versions of environments. For PyBullet we used \texttt{v0} versions of environments.

For AdaTQC and AdaMMQL we have used original authors implementations of TQC\footnote{\href{https://github.com/SamsungLabs/tqc\_pytorch}{https://github.com/SamsungLabs/tqc\_pytorch}} and MMQL\footnote{\href{https://github.com/qlan3/Explorer}{https://github.com/qlan3/Explorer}}. Authors of WD3 do not provide their code so we reimplement their method based on original code of TQC.

\begin{figure*}[!htb]
\centering
\includegraphics[width=\textwidth]{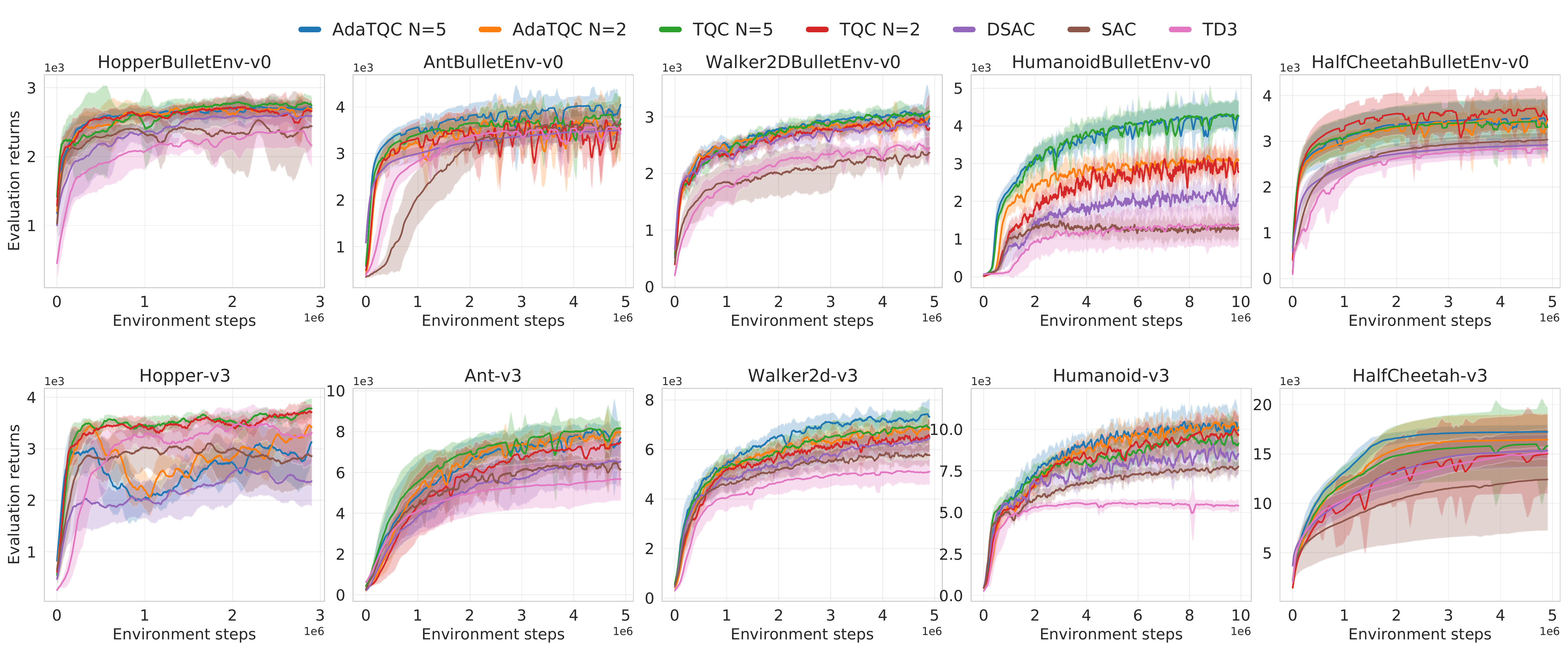}
\caption{Learning curves for AdaTQC, TQC and a number of baselines: DSAC \cite{ma2020dsac}, SAC \cite{haarnoja2018softapp}, TD3 \cite{fujimoto2018addressing}. We can see, that except 'Hopper-v3' environment, AdaTQC repeats the curve and performance of TQC with the optimal parameter, and on some environments ('Walker2d-v3', for example) even outperforms TQC. Each curve consist of 10 averaged runs, performance is evaluated each 1000 steps, curves are smoothed with window 100, $\pm$ standard deviation of 10 runs is shaded. 
}
\label{fig:main_baselines}
\end{figure*}

In the original paper \cite{kuznetsov2020controlling} the core hyperparameter is the number of dropped quantiles $d$ per critic. In terms of this paper $\eta^\text{TQC} = N \cdot d = 2\cdot d$, where $N$ is the number of critics. For Figure~\ref{fig:main_baselines} we take optimal values of $\eta^\text{TQC}$ for TQC on MuJoCo from the original paper. For PyBullet we have test $6$ different values and selected the optimal one based on $4$ runs with each hyperparameter.

We list all the hyperparameters in Table~\ref{tab:hyp_tqc} for TQC and AdaTQC, in Table~\ref{tab:hyp_wd3} for WD3 and AdaWD3, and in Table~\ref{tab:hyp_maxmin} for \maxmin~and Ada\maxmin.

\begin{table*}[t!]
\centering
\resizebox{0.6\textwidth}{!}{
\begin{tabular}{lccccc}
\toprule
\bf PyBullet Environment & Humanoid & Walker2D & Hopper & HalfCheetah & Ant\\
\midrule
\bf Optimal $d$ & 3        & 5        & 5      & 2           & 3\\
\bf Optimal $\eta^\text{TQC}$ & 6        & 10        & 10      & 4           & 6\\
\bottomrule
\end{tabular}
}
\caption{
Optimal values for the number of truncated atoms $d$/$\eta^\text{TQC}$ for TQC on PyBullet.
}
\label{tab:optimal_bullet}
\end{table*}

\begin{table}[t]
\label{tab:hyper_tqc}
\begin{center}
\begin{small}
\begin{tabular}{lcc}
\toprule
Hyperparameter & AdaTQC & TQC \\
\midrule
Optimizer & \multicolumn{2}{c}{Adam} \\
Learning rate & \multicolumn{2}{c}{\num{3e-4}} \\
Discount $\gamma$ & \multicolumn{2}{c}{0.99} \\
Replay buffer size & \multicolumn{2}{c}{\num{1e6}} \\
Number of critics $N$ & \multicolumn{2}{c}{2} \\
Number of hidden layers in critic networks & \multicolumn{2}{c}{3} \\
Size of hidden layers in critic networks & \multicolumn{2}{c}{512} \\
Number of hidden layers in policy network & \multicolumn{2}{c}{2} \\
Size of hidden layers in policy network & \multicolumn{2}{c}{256} \\
Minibatch size & \multicolumn{2}{c}{256} \\
Entropy target $\mathcal{H}_T$ &  \multicolumn{2}{c}{$- \dim \mathcal{A}$} \\
Tempereture parameter $\alpha$ & \multicolumn{2}{c}{auto}\\
Nonlinearity & \multicolumn{2}{c}{ReLU} \\
Target smoothing coefficient $\tau$ & \multicolumn{2}{c}{0.005} \\
Target update interval & \multicolumn{2}{c}{1} \\
Gradient steps per iteration & \multicolumn{2}{c}{1} \\
Environment steps per iteration & \multicolumn{2}{c}{1} \\
Number of atoms $M$ & \multicolumn{2}{c}{25} \\
Huber loss parameter $\kappa$ & \multicolumn{2}{c}{1} \\
\midrule
Bias evaluation period $M_\text{compute}$ & 10 & -- \\
Bias averaging coefficient $\gamma_\eta$ & 0.999 & -- \\
Bias update interval $M_\text{update}$ & 50000 & -- \\
Replay buffer $\mathcal{D}_\text{fresh}$ size in trajectories $N_\text{fresh}$ & 200 & -- \\
Batch size from the fresh buffer $\mathcal{D}_\text{fresh}$ $S_\text{fresh}$ & 4000 & -- \\
Rollout length $k$ & 500 & -- \\
\bottomrule
\end{tabular}
\end{small}
\caption{Hyperparameter values for TQC and AdaTQC.}
\label{tab:hyp_tqc}
\end{center}
\end{table}

\begin{table}[t]
\begin{center}
\begin{small}
\begin{tabular}{lcc}
\toprule
Hyperparameter & AdaWD3 & WD3 \\
\midrule
Optimizer & \multicolumn{2}{c}{Adam} \\
Learning rate & \multicolumn{2}{c}{\num{3e-4}} \\
Discount $\gamma$ & \multicolumn{2}{c}{0.99} \\
Replay buffer size & \multicolumn{2}{c}{\num{1e6}} \\
Number of critics $N$ & \multicolumn{2}{c}{2} \\
Number of hidden layers in critic networks & \multicolumn{2}{c}{2} \\
Size of hidden layers in critic networks & \multicolumn{2}{c}{256} \\
Number of hidden layers in policy network & \multicolumn{2}{c}{2} \\
Size of hidden layers in policy network & \multicolumn{2}{c}{256} \\
Minibatch size & \multicolumn{2}{c}{100} \\
Nonlinearity & \multicolumn{2}{c}{ReLU} \\
Target smoothing coefficient $\tau$ & \multicolumn{2}{c}{0.005} \\
Target update interval & \multicolumn{2}{c}{2} \\
Policy update interval & \multicolumn{2}{c}{2} \\
Exploration noise standard deviation & \multicolumn{2}{c}{0.1} \\
Standard deviation of policy noise for TD update   & \multicolumn{2}{c}{0.2} \\
Clipping values for TD update policy noise & \multicolumn{2}{c}{$[-0.5, 0.5]$} \\
\midrule
Bias evaluation period $M_\text{compute}$ & 10 & -- \\
$\eta$ learning rate $\lambda$ & 3e-5 & -- \\
Replay buffer $\mathcal{D}_\text{fresh}$ size in trajectories $N_\text{fresh}$ & 200 & -- \\
Batch size from the fresh buffer $\mathcal{D}_\text{fresh}$ $S_\text{fresh}$ & 4000 & -- \\
Rollout length $k$ & 500 & -- \\
\bottomrule
\end{tabular}
\end{small}
\caption{Hyperparameter values for WD3 and AdaWD3.}
\label{tab:hyp_wd3}
\end{center}
\end{table}

\begin{table}[t]
\label{tab:hyper_mmql}
\begin{center}
\begin{small}
\begin{tabular}{lcccc}
\toprule
& \multicolumn{2}{c}{MinAtar} & \multicolumn{2}{c}{Pixelcopter} \\
Hyperparameter & AdaMMQL & MMQL & AdaMMQL & MMQL \\
\midrule
Max episode length & \multicolumn{2}{c}{\num{1e+4}} & \multicolumn{2}{c}{500}\\
Output type of environment & \multicolumn{2}{c}{\textit{pixel}} & \multicolumn{2}{c}{\textit{feature}}\\
Number of convolutional layers & \multicolumn{2}{c}{1} & \multicolumn{2}{c}{--}\\
Number of input channels & \multicolumn{2}{c}{4} & \multicolumn{2}{c}{--}\\
Number of output channels & \multicolumn{2}{c}{16} & \multicolumn{2}{c}{--}\\
Kernel size & \multicolumn{2}{c}{4} & \multicolumn{2}{c}{--}\\
Stride & \multicolumn{2}{c}{1} & \multicolumn{2}{c}{--}\\
Sizes of hidden layers & \multicolumn{2}{c}{$[128]$} &\multicolumn{2}{c}{$[64, 64]$}\\
Total amount of networks $N_{tot}$ &\multicolumn{4}{c}{8}\\
Number of updated networks $N_{upd}$ & \multicolumn{4}{c}{2}\\
Optimizer & \multicolumn{4}{c}{RMSprop} \\
Learning rate & \multicolumn{4}{c}{\num{3e-4}} \\
Optimizer $\alpha$ & \multicolumn{2}{c}{0.95} & \multicolumn{2}{c}{0.99}\\
Optimizer $\epsilon$ & \multicolumn{2}{c}{0.01} & \multicolumn{2}{c}{\num{1e-8}}\\
Optimizer \textit{centered} & \multicolumn{2}{c}{\textit{true}} & \multicolumn{2}{c}{\textit{false}} \\
Replay buffer size & \multicolumn{4}{c}{\num{1e+5}}\\
Exploration steps & \multicolumn{2}{c}{\num{5e+3}} & \multicolumn{2}{c}{\num{1e+3}}\\
$\epsilon_{steps}$ & \multicolumn{2}{c}{\num{1e+5}} & \multicolumn{2}{c}{\num{1e+3}}\\
$\epsilon_{start}$ & \multicolumn{4}{c}{1}\\
$\epsilon_{end}$ & \multicolumn{2}{c}{0.1} & \multicolumn{2}{c}{0.01}\\
$\epsilon_{decay}$ & \multicolumn{4}{c}{0.999}\\
Loss for Q-Learning & \multicolumn{2}{c}{\textit{SmoothL1Loss}} & \multicolumn{2}{c}{\textit{MSELoss}}\\
Batch size from replay buffer & \multicolumn{4}{c}{32}\\
Discount $\gamma$ & \multicolumn{4}{c}{0.99}\\
Target network update frequency & \multicolumn{4}{c}{1000}\\
Max norm for gradient clipping & \multicolumn{2}{c}{\textit{None}}& \multicolumn{2}{c}{5}\\
\midrule
Bias evaluation period $M_\text{compute}$ & 10 & --  & 10 & --  \\
Bias averaging coefficient $\gamma_\eta$ & 0.999 & -- & 0.999 & --\\
Bias update interval $M_\text{update}$ & 50000 & -- & 50000 & -- \\
Replay buffer $\mathcal{D}_\text{fresh}$ size in trajectories $N_\text{fresh}$ & 200 & -- & 200 & --\\
Batch size from the fresh buffer $\mathcal{D}_\text{fresh}$ $S_\text{fresh}$ & 4000 & -- & 4000 & --\\
Rollout length $k$ & 200 & -- & 200 & --\\
\bottomrule
\end{tabular}
\end{small}
\caption{Hyperparameter values for MMQL and AdaMMQL. All hyperparameters non-related to the adaptive version of the method except for $N_{tot}$ and $N_{upd}$ were taken from the paper~\citep{lan2020maxmin}.}
\label{tab:hyp_maxmin}
\end{center}
\end{table}

\section{Modification of MMQL}
\label{app:mmql_mod}
While experimenting with Maxmin Q-learning we found that one implementation detail severely affects the performance while being disconnected from the core idea.
Specifically, in the original implementation of \maxmin~only one out of $\eta^{\sl \text{\maxmin}}$ random  Q-network is updated at each step to keep computational costs constant with respect to $\eta^{\sl \text{\maxmin}}$.
We found that the dependence of control hyperparameter $\eta^{\sl \text{\maxmin}}$ on this update rate ($\frac{1}{\eta^{\sl \text{\maxmin}}}$) hinders the impact of the overestimation control on the bias and performance.
To decouple the effect of overestimation control from the impact of update rate we keep the update rate constant.

Moreover we decided to balance the ensembling effect naturally arising in Ada\maxmin --- it must be able to use more critics than the current value of $\eta^\maxmin$ since it should be ready to increase  $\eta^\maxmin$ --- by introducing the similar ensembling scheme into \maxmin.

Summing up the two above modifications for both \maxmin~and Ada\maxmin~we:
\begin{itemize}
    \item fix some reasonably big total number of Q-networks $N_{tot}=8$
    \item fix the update rate for all Q-networks to $0.25=\frac{N_{upd}}{N_{tot}}=\frac{2}{N_{tot}}$
    \item on each step sample without repetition $\eta^{\sl \text{\maxmin}}$ out of $N_{tot}=8$ Q-networks from  Q-networks to calculate the target and select the action
\end{itemize}

That way we can properly estimate the effect of overestimation control on the performance and fairly compare adaptive and non-adaptive methods since for any specific value of $\eta^{\sl \text{\maxmin}}$ adaptive method works exactly the same as non-adaptive one.

\section{Sample efficiency computation}
\label{app:sample_eff_comp}
In this section, we describe how we quantified sample efficiency.
Sample inefficiency of non-adaptive methods come from the need for an environment-specific search for the number of truncated quantities $d$.
In Table~\ref{table:sample_eff_tqc} we compared the number of samples required by adaptive method to that of a conventional one with a grid search. 

Column ``ISE'' is calculated as follows:
\begin{enumerate}
    \item For each environment and method (non-adaptive method with some $\eta$ from grid $G$, and adaptive one), we consider the average of evaluation performance over the last $10^5$ interactions and over $4$ seeds as the \textit{final performance}.
    \item For a fixed number $n \in [1, \dots, |G|]$ of different values for hyperparameter $\eta \in G$ we average over all possible combinations~(without replacement) the maximum of \textit{final performance} within the combination ($n$ out of $|G|$) and call it \textit{hp-averaged performance}.
    \item For each environment we find the least $n$ such that the corresponding \textit{hp-averaged performance} is less than the \textit{final performance} of adaptive method and call it \textit{ISE}. 
\end{enumerate}

In other words, ``ISE'' is the minimum amount of hyperparameter tries $n$, that on average is required by non-adaptive method to achieve the same or greater \textit{final performance} than adaptive method.

\section{On the optimization of continuous \texorpdfstring{$\eta$}{eta}}
Update rule provided in equation~\ref{eq:eta-update} is realized in practice via gradient steps with Adam optimizer \citep{kingma2015adam} on the following loss:

\begin{equation}
    \eta \left(\wavyaggbias(\hat{Q}_{\psi}, \mathcal{B})\right)^- \rightarrow \max_\eta
\end{equation}

where $(\dots)^-$ means stop-gradient.
Because Adam normalizes the gradients, in practice we get an optimization process where all absolute differences between neighboring values of $\eta$ are almost the same. 

To not confuse the reader that the magnitude of bias matters in this optimization process, we have decided to write $\eta$ optimization in form \ref{eq:eta-update}.
\pagebreak

\section{AdaTQC and TQC overestimation measurements}
\label{sec:bias_eval_app}
\begin{figure*}[!htb]
\centering
\includegraphics[width=\textwidth]{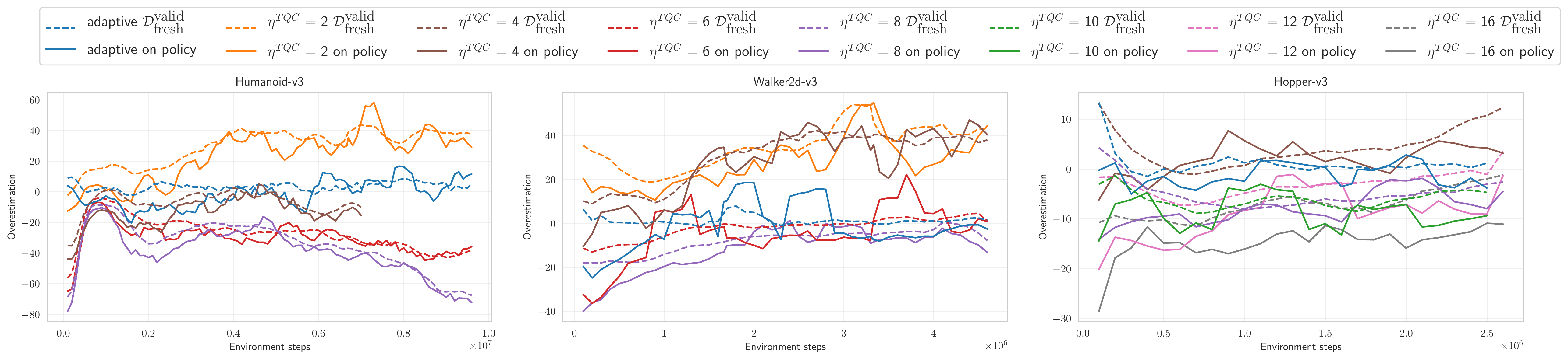}
\caption{
Unbiased estimate of the aggregated bias using additional data (on policy) and biased estimate based on $\mathcal{D}^{\text{valid}}_{\text{fresh}}$ (used by adaptive methods and not used by non-adaptive methods) for TQC and AdaTQC.
 $\mathcal{D}^{\text{valid}}_{\text{fresh}}$-based estimate of the aggregated bias closely matches the unbiased on-policy version of the estimate. With adaptive control of $\eta^{TQC}$ the bias remains close to zero, confirming that AdaTQC is indeed able to control the bias. Data points are calculated each 100k steps, the curves are averaged over 4 runs and smoothed with window of 5.
}
\label{fig:bias_eval_app}
\end{figure*}

\FloatBarrier

\section{Learning curves}
\label{sec:learning_curves}

\subsection{AdaTQC, TQC and baselines}
\FloatBarrier
\label{sec:adatqc_curves}

\begin{figure*}[!htb]
\centering
\includegraphics[width=\textwidth]{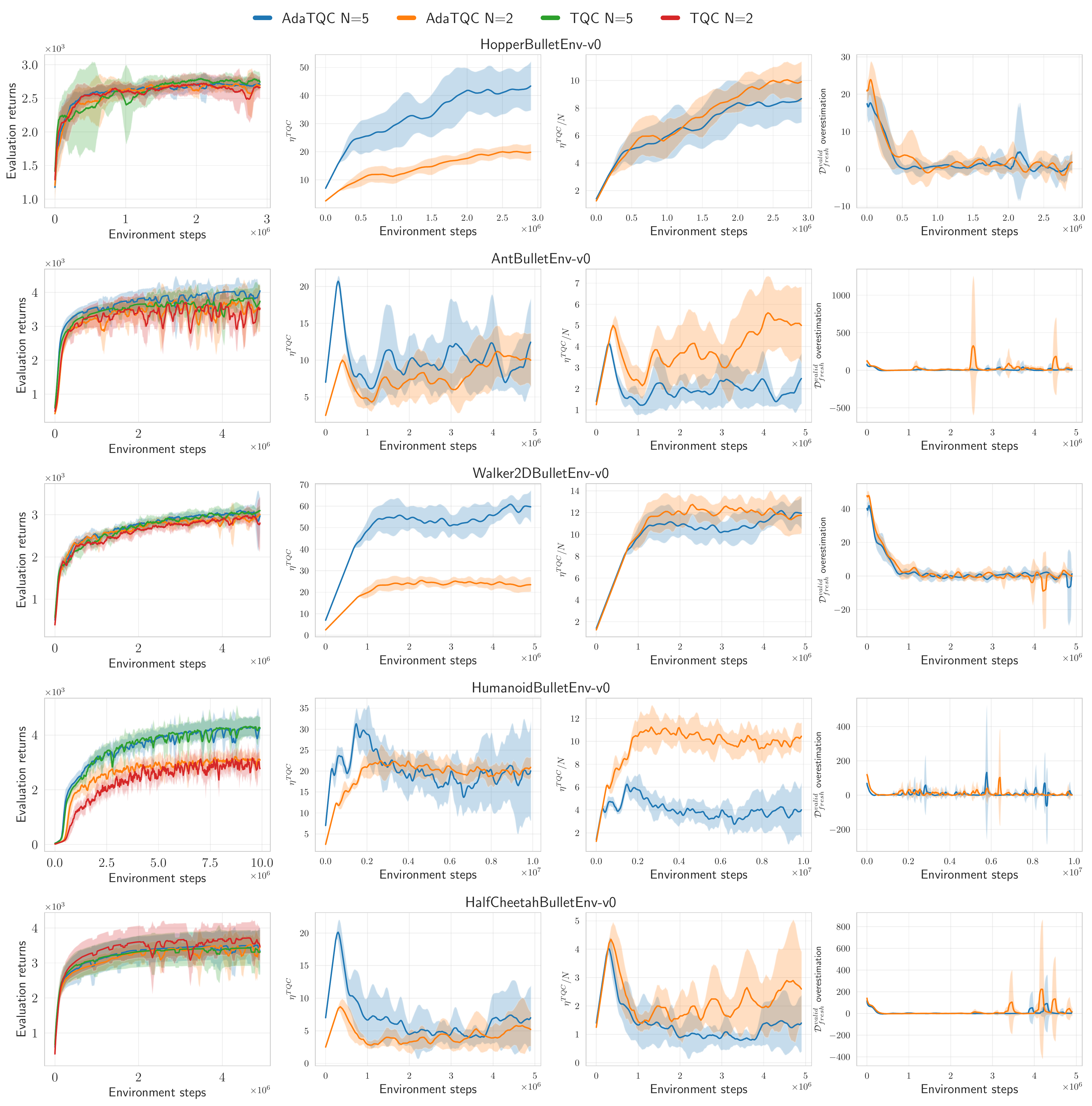}
\caption{Learning curves for TQC and AdaTQC with number of critics $N=2,5$ on PyBullet environments. Next figure contains results on MuJoCo environments and conclusions.
}
\label{fig:tqc_app}
\end{figure*}

\begin{figure*}[!htb]
\centering
\includegraphics[width=\textwidth]{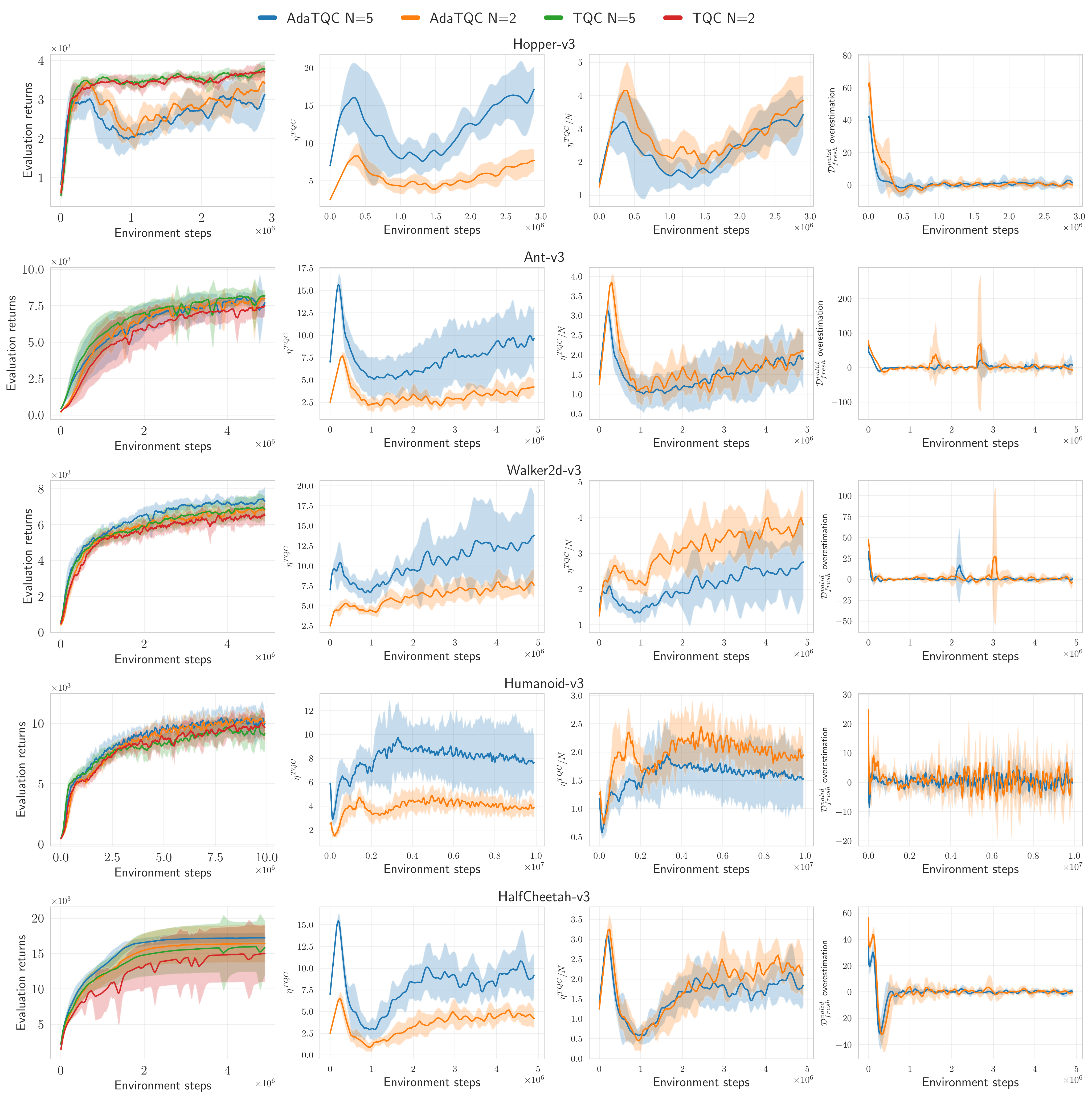}
\caption{
Learning curves for TQC and AdaTQC with number of critics $N=2,5$ on MuJoCo environments. We can see, that number of dropped quantiles per network $\eta^{TQC}/N$ is lower for $N=5$. It means that increasing the ensemble size reduces overestimation by itself, and lesser amount of external influence is needed. Each curve consist of 10 averaged runs, performance is evaluated each 1000 steps, curves are smoothed with window 100, $\pm$ standard deviation of 10 runs is shaded. 
}
\label{fig:tqc_app_2}
\end{figure*}

\FloatBarrier

\subsection{AdaWD3 and WD3}
\begin{figure*}[!htb]
\centering
\includegraphics[width=\textwidth]{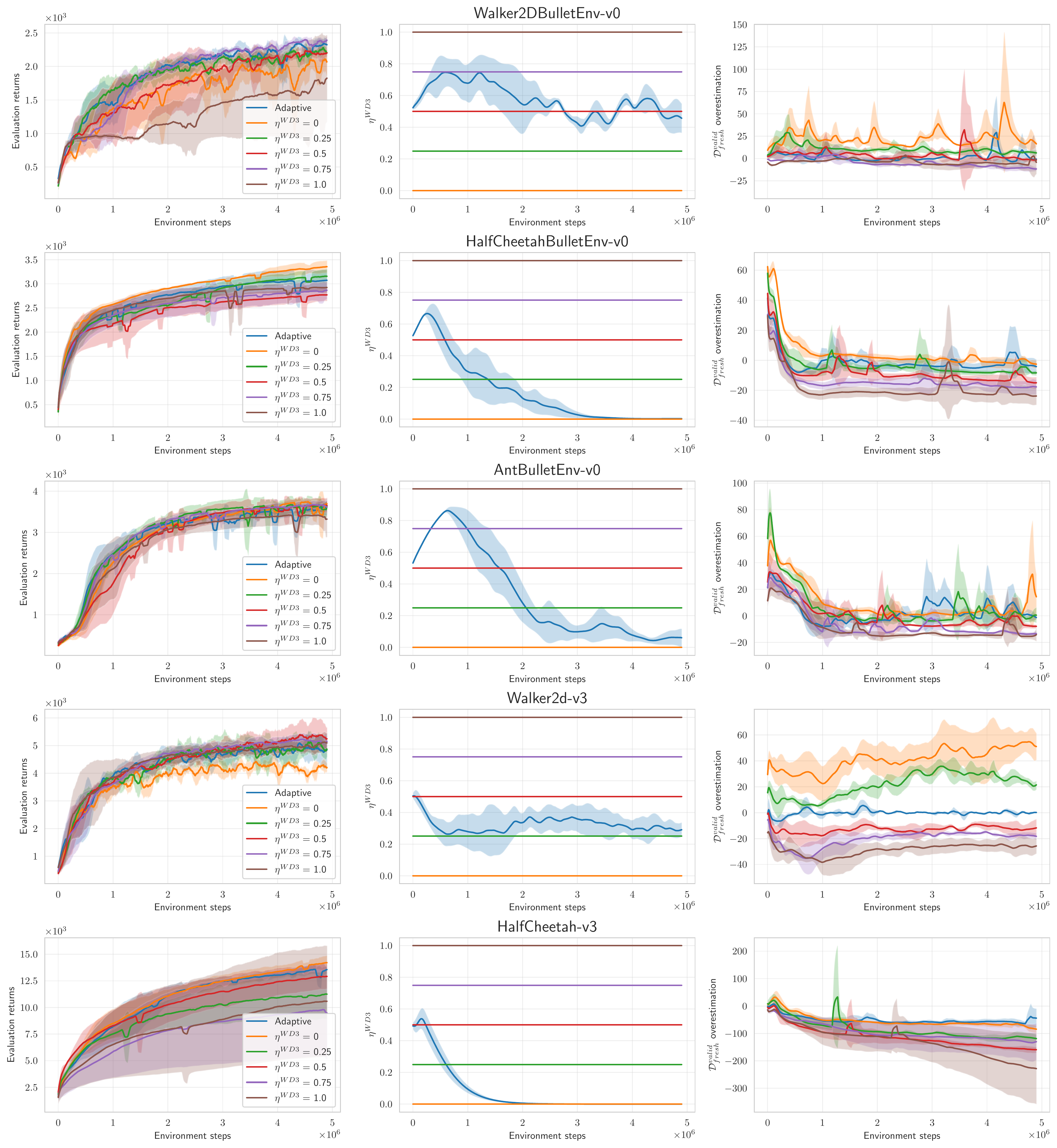}
\caption{
Learning curves for WD3 and AdaWD3 on subset of MuJoCo and PyBullet environments. We can see, that in most cases automatic tuning is able to control estimated bias. One exception is 'HalfCheetah-v3', where the original method is not flexible enough. Each curve consist of 4 averaged runs, performance is evaluated each 1000 steps, curves are smoothed with window 100, $\pm$ standard deviation of 4 runs is shaded. 
}
\label{fig:wd3_app}
\end{figure*}

\FloatBarrier
\pagebreak
\subsection{MMQL and AdaMMQL}
\begin{figure*}[!htb]
\centering
\includegraphics[width=\textwidth]{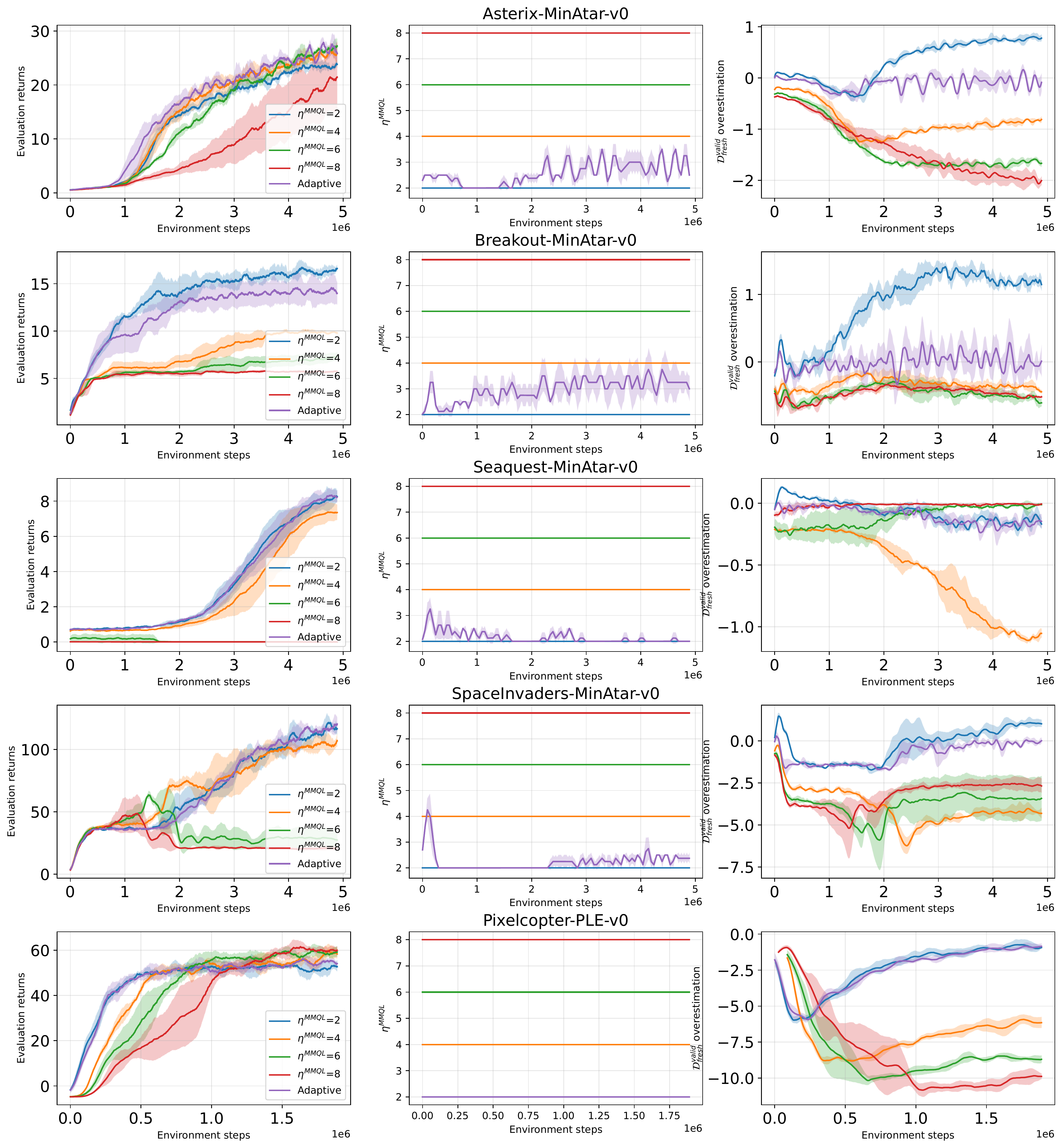}
\caption{
Learning curves for \maxmin~and Ada\maxmin~on subset of MinAtar and PLE environments. We can see, that in most cases automatic tuning is able to control estimated bias. One exception is 'Pixelcopter-PLE-v0', where the original method is not flexible enough. Each curve consist of 4 averaged runs, performance is evaluated each 1000 steps, curves are smoothed with window 100, $\pm$ standard deviation of 4 runs is shaded. 
}
\label{fig:maxmin_app}
\end{figure*}
\end{appendices}
\end{document}